\documentclass{article} % For LaTeX2e
\usepackage[preprint]{colm2026_conference}
\usepackage{microtype}
\usepackage{url}
\usepackage{booktabs}

\usepackage{amsmath}
\usepackage{mathtools}
\usepackage{amsthm}
\usepackage{lineno}
\usepackage{graphicx}
\usepackage{amssymb}
\usepackage{hyperref}
\usepackage{pifont}
\usepackage{xspace}
\usepackage{caption}
\usepackage{multirow}
\usepackage{colortbl}
\usepackage{color}
\usepackage{enumitem}
\usepackage{tcolorbox}
\tcbuselibrary{breakable}

\theoremstyle{plain}

\theoremstyle{definition}

\theoremstyle{remark}

\definecolor{beigecolor}{RGB}{253, 244, 204} % Beige color
\definecolor{greencolor}{RGB}{228, 242, 217} % Green color
\definecolor{bluecolor}{RGB}{66, 133, 244} % Beige color
\definecolor{orgcolor}{RGB}{255, 140, 15} % Green color
\definecolor{redcolor}{RGB}{234, 67, 53} % Green color
\definecolor{ggreen}{RGB}{52, 168, 83}
\definecolor{gyellow}{RGB}{251, 188, 5}
\definecolor{lightorange}{RGB}{245, 237, 211} % Adjust these values as needed
\definecolor{bluebar}{RGB}{138,159,201}
\definecolor{pinkbar}{RGB}{232,180,189}

\usepackage[textsize=tiny]{todonotes}
\usepackage{mathtools}

\definecolor{darkblue}{rgb}{0, 0, 0.5}
\hypersetup{colorlinks=true, citecolor=darkblue, linkcolor=darkblue, urlcolor=darkblue}

\title{Escaping the Context Bottleneck: Active Context Curation for LLM Agents via Reinforcement Learning}

\author{Xiaozhe Li $^{1}$ $^{*}$, Tianyi Lyu $^{1}$ $^{*}$, Yizhao Yang $^{1}$, Liang Shan $^{1}$, Siyi Yang $^{1}$, Ligao Zhang $^{3}$ \\ \textbf{Zhuoyi Huang $^{2}$ $^{3}$}, \textbf{Qingwen Liu $^{1}$}, \textbf{Yang Li $^{4}$}\\
$^{1}$ Tongji University, $^{2}$Stanford University, $^{3}$CurrentsAI Research, $^{4}$ Independent\\
}

\begin{document}

\ifcolmsubmission
\linenumbers
\fi

\maketitle

\begin{abstract}
Large Language Models (LLMs) struggle with long-horizon tasks due to the "Context Bottleneck" and the "Lost-in-the-Middle" phenomenon, where accumulated noise from verbose environments degrades reasoning over multi-turn interactions. To address this, we introduce a symbiotic framework that decouples context management from task execution. Our architecture pairs a lightweight, specialized policy model (ContextCurator) that actively curates the working memory for a powerful, frozen foundation model (TaskExecutor). Trained via reinforcement learning, the ContextCurator performs active information entropy reduction. It aggressively prunes environmental noise while meticulously preserving "reasoning anchors"—sparse data points critical for future deductions. On WebArena, our framework boosts Gemini-3.0-flash's success rate from 36.4\% to 41.2\% while reducing token consumption by 8.8\% ($47.4K \rightarrow 43.3K$). On DeepSearch, it achieves 57.1\% success (vs. 53.9\%) with an $8\times$ token reduction. Remarkably, a 7B ContextCurator matches the context management performance of GPT-4o, providing a scalable, computationally efficient paradigm for autonomous long-horizon agents.
\end{abstract}

\section{Introduction}\label{sec:intro}

The trajectory of artificial intelligence has fundamentally shifted from the development of static conversational models to the deployment of autonomous, goal-oriented Large Language Model (LLM) agents capable of sequential decision-making in complex environments~\citep{wang2024survey, liu2025advances, grpo, o1}. This advancement enables agents to tackle complex real-world challenges, from web browsing~\citep{gur2024a, yao2022webshop} to multi-hop information retrieval~\citep{yang2024sweagent, xie2024osworld}. However, a critical bottleneck emerges in long-horizon tasks: while frontier models exhibit profound zero-shot reasoning capabilities, their performance degrades dramatically due to the inherent limitations of their \textit{cognitive span}~\citep{huang2025language, du2025rethinking}.

The fundamental challenge to reliable long-horizon execution is the ability to maintain task-relevant working memory while systematically filtering environmental noise over extended trajectories. Real-world observation spaces are characterized by extreme informational entropy. In web tasks, DOM trees contain over 90\% structural noise (e.g., advertisements, styling scripts, redundant navigational elements), while search paradigms flood the context with semantic noise and distractor paragraphs. When agents process these raw streams, the signal-to-noise ratio (SNR) deteriorates rapidly. Feeding this high-entropy data directly into a transformer inevitably triggers the \textit{"Lost-in-the-Middle" phenomenon} and causes \textit{cascading logical failures}, where early noise compounds into hallucinated reasoning in subsequent steps. Our experiments reveal that even frontier models like Gemini-3.0-flash achieve only 36.4\% success on WebArena and 53.9\% on DeepSearch (Tables~\ref{tab:webarena} and~\ref{tab:search}). These results expose a critical limitation: brute-force expansion of context windows does not guarantee effective reasoning. The problem is not capacity, but signal-to-noise ratio.

Current mitigation strategies fall short of addressing the dynamic nature of these environments. \textit{Passive memory systems}~\citep{chhikara2025mem0,packer2023memgpt,zhong2024memorybank,reasoningbank} treat context management as a static retrieval problem, relying on semantic similarity to recall information. These methods suffer from retrieval bias; they frequently fail to retrieve implicit \textbf{reasoning anchors}—causally essential information that may be textually dissimilar to the current query. Conversely, \textit{monolithic learned architectures} attempt to internalize memory management within the same task model~\citep{zhou2025mem1,yu2025memagent}. This creates a severe capacity split: small models lack the reasoning depth required for complex logical execution, while large proprietary models are too computationally expensive and opaque for direct on-policy fine-tuning.

To bridge this gap, we introduce \textbf{ActiveContext}, a framework that reframes memory management from a static storage utility into an \textbf{active, sequential decision-making process}. We adopt a complementary cognitive architecture that decouples the system into two specialized components: a powerful frozen foundation model (\textbf{TaskExecutor}) handles pure reasoning and task execution, while a lightweight specialized policy model (\textbf{ContextCurator}) actively curates its working memory.

Trained via reinforcement learning, the ContextCurator learns to act as a "Cognitive Architect." Driven by a distal reward signal based on the TaskExecutor's success, it develops a generalizable policy to perform active \textbf{information entropy reduction}. It aggressively prunes structural and semantic noise while meticulously preserving \textbf{reasoning anchors}—sparse data points critical for future deductions. This symbiotic design ensures that the TaskExecutor's attention heads are focused solely on high-fidelity, high-signal data.

Extensive evaluation demonstrates that our decoupled framework establishes a new Pareto frontier in agentic reasoning. On WebArena, ActiveContext boosts Gemini-3.0-flash from a 36.4\% to a 41.2\% success rate while reducing token consumption by 8.8\% (47.4K $\rightarrow$ 43.3K). On DeepSearch, the success rate improves from 53.9\% to 57.1\% with nearly an $8\times$ token reduction (46.7K $\rightarrow$ 6.6K). Remarkably, our 7B ContextCurator matches the context management performance of GPT-4o. This result democratizes agentic pipelines, proving that active memory curation is a distinct cognitive skill that can be efficiently offloaded to smaller models.

Our work makes three key contributions:
\begin{itemize}[leftmargin=*, topsep=0pt, itemsep=0pt]
\item We propose a novel \textbf{symbiotic framework} that decouples context management from task execution. By pairing a specialized ContextCurator with a closed-source SOTA TaskExecutor, we transform memory from a passive storage utility into an active, decision-dependent control variable.

\item We formulate an \textbf{on-policy RL pipeline} that trains lightweight models to actively manage context through dynamic environment interaction. Guided by distal rewards, the model learns a generalizable policy for \textbf{information entropy reduction}—aggressively pruning noise while meticulously preserving critical reasoning anchors.

\item We demonstrate substantial empirical gains across complex long-horizon benchmarks. Our framework improves SOTA model performance while reducing token consumption, and proves that a specialized 7B model can match the context management capabilities of GPT-4o, providing a highly scalable and sustainable paradigm for autonomous agents.
\end{itemize}

\vspace{-5pt}
\section{Related Work}\label{sec:related_work}
\vspace{-5pt}

\textbf{Memory for Agentic Systems.}
Context memory enables autonomous agents to maintain state and leverage history for complex reasoning~\citep{zhang2024survey}. Existing frameworks typically utilize passive storage (e.g., Mem0~\citep{chhikara2025mem0}, MemGPT~\citep{packer2023memgpt}, LangMem\citep{langgraph}) via semantic retrieval, or procedural workflows (e.g., HiAgent~\citep{hu-etal-2025-hiagent}, AWM~\citep{wang2025agent}) to induce structured trajectories. However, passive systems suffer from retrieval bias, frequently missing causally essential "reasoning anchors" that lack textual similarity~\citep{ming2025anchorcot,zhou2024webarena}, while procedural workflows struggle to generalize across high-entropy domains. Advanced cognitive graphs (e.g., Synapse~\citep{zheng2024synapse}, MAGMA~\citep{jiang2026magma}) mitigate this but introduce formidable computational latency. In contrast, our framework adopts a \textbf{symbiotic architecture} that decouples \textit{context management} from \textit{task execution}. Rather than relying on static storage or heavy graphs, our \textit{ContextCurator} actively curates context to maximize real-time information density.

\textbf{Reinforcement Learning for LLM Agents.}
RL is a pivotal paradigm for aligning agentic reasoning across diverse domains (e.g., web navigation, embodied environments) using algorithms like PPO and GRPO~\citep{ppo,grpo,yao2022webshop,webagentr1,searchr1}. Recognizing the limits of static memory, recent works (e.g., Mem1~\citep{zhou2025mem1}, MemAgent~\citep{yu2025memagent}, MemoryR1~\citep{yan2025memory}) apply RL to optimize active memory operations as sequential decisions. However, these approaches train a \textit{monolithic} model for both memory curation and task execution. This creates a severe capacity split: small open-source models lack the deep reasoning required for complex deductions, while frontier models are opaque and computationally prohibitive for on-policy fine-tuning. Monolithic RL also risks inducing "lazy agent" shortcut behaviors~\citep{zhang2025unlocking}. \textit{ActiveContext} deviates via a \textbf{collaborative symbiotic architecture}. We use RL to train a lightweight \textit{ContextCurator} that strictly manages context for a powerful, frozen \textit{TaskExecutor}, combining adaptive entropy reduction with frontier reasoning strength at a fraction of the cost.

\begin{figure*}[t]
    \centering
    \includegraphics[width=\linewidth]{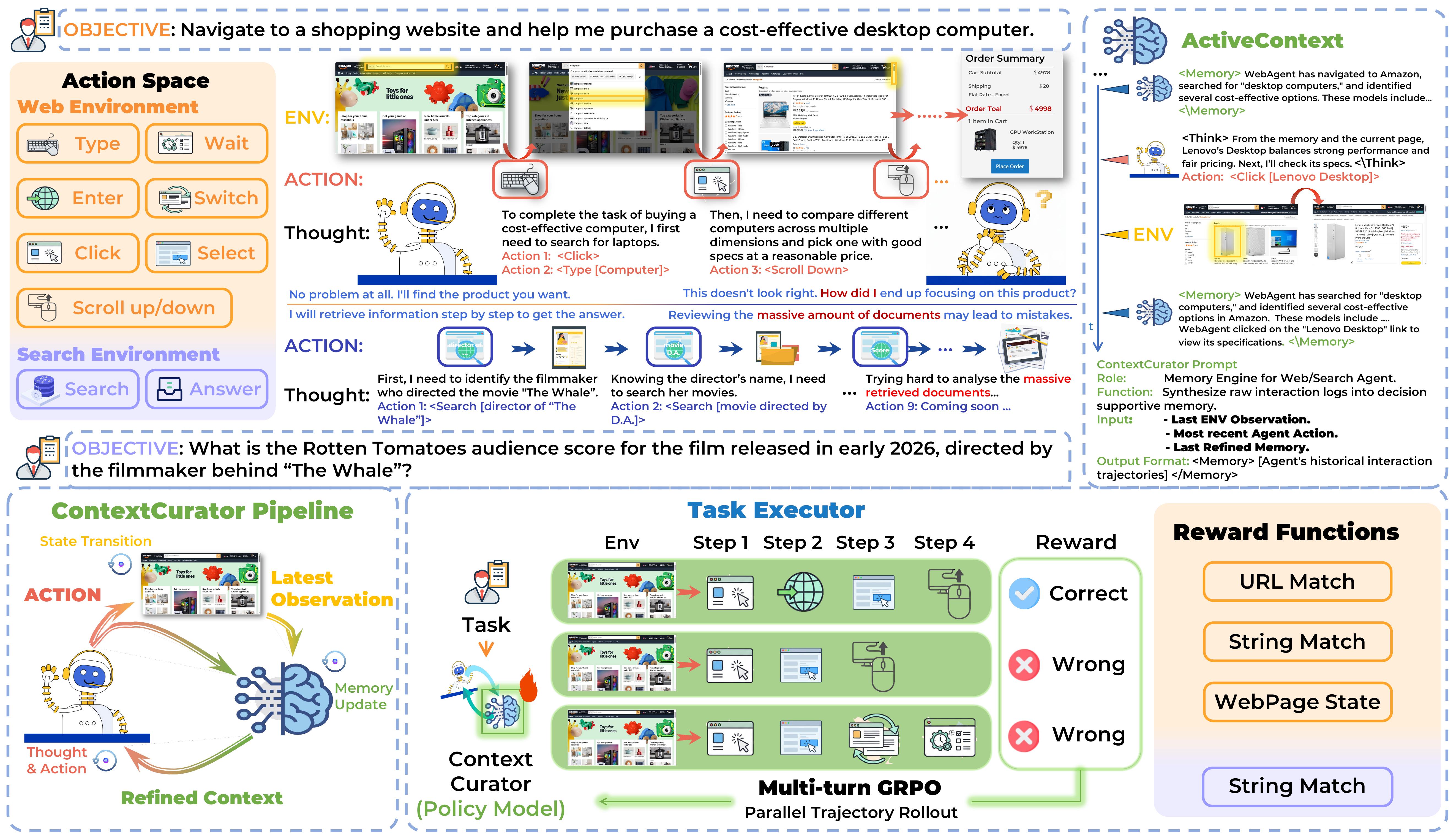}
    % \caption{Overview of the \textbf{ActiveContext} framework for web and search environments. The top-left panel shows the interaction of a single LLM agent in both web and search settings. In the web environment, the agent performs actions like typing, clicking, and scrolling to achieve its objective. In the search environment, the agent utilizes internal knowledge or a retrieval database to answer questions. The central section highlights the agent's reasoning process, addressing challenges like context bottleneck and the "lost-in-the-middle" phenomenon. The top-right panel demonstrates the active context management strategies, where \textit{ContextCurator} refines context and guides action selection. The bottom-left panel illustrates the input-output flow of the ContextCurator, showing the agent-environment interaction at each step. The bottom-right panel depicts \textit{Multi-turn GRPO} for sequential decision-making, where the agent rolls out parallel trajectories to collect rewards and refine its context policy for improved task success.}
    \caption{Overview of \textbf{ActiveContext}. \textbf{Top/Center:} Standard LLM agent interactions in web and search domains, illustrating the context bottleneck and "lost-in-the-middle" phenomenon during multi-turn reasoning. \textbf{Top-Right \& Bottom-Left:} Our symbiotic architecture, where the \textit{ContextCurator} actively refines verbose observations into a high-fidelity working memory for the \textit{TaskExecutor}. \textbf{Bottom-Right:} The \textit{Multi-turn GRPO} pipeline, which rolls out trajectories to collect distal rewards and optimize the context curation policy.}
    \label{fig:pipeline}
    \vspace{-10pt}
\end{figure*}

\vspace{-5pt}
\section{Methodology}\label{sec:methodology}
\vspace{-3pt}

% We propose \textbf{ActiveContext}, a symbiotic agentic framework designed to resolve the "Context Bottleneck" inherent in long-horizon tasks. Conventional approaches, such as Mem0~\citep{chhikara2025mem0}, MemoryBank~\citep{zhong2024memorybank} and ReasoningBank~\citep{reasoningbank}, typically treat context management as a static retrieval problem. They rely on heuristic pipelines to extract structured memory items and use embedding-based similarity to recall them. While interpretable, these methods suffer from two critical limitations when applied to high-entropy dynamic environments: (1) \textbf{Passivity and Rigidity:} they rely on brittle, hand-crafted workflows that lack the adaptability to discern nuanced, task-specific signals; and (2) \textbf{Retrieval Bias:} semantic similarity search often fails to retrieve "reasoning anchors"—causally essential information that may lack textual similarity to the current query. To transcend these limitations, ActiveContext reframes context management not as a storage and retrieval task, but as an \textit{active, sequential decision-making problem}.

We propose \textbf{ActiveContext}, a symbiotic framework designed to resolve the "Context Bottleneck" in long-horizon tasks. Conventional approaches (e.g., Mem0~\citep{chhikara2025mem0}, MemoryBank~\citep{zhong2024memorybank}, ReasoningBank~\citep{reasoningbank}) treat memory as a static retrieval problem. In high-entropy dynamic environments, these methods suffer from \textbf{rigidity} via brittle, hand-crafted workflows, and \textbf{retrieval bias}—semantic searches frequently fail to recall causally essential "reasoning anchors" lacking textual similarity to the current query. To transcend these limitations, ActiveContext reframes context management from a passive storage utility into an \textit{active, sequential decision-making problem}.

\subsection{Problem Formulation}\label{sec:problem_formulation}

% We formulate the interaction between an autonomous LLM agent and a complex, long-horizon environment (e.g., a web browser or search engine) as a Partially Observable Markov Decision Process (POMDP), defined by the tuple $\mathcal{M} = (\mathcal{S}, \mathcal{A}, \mathcal{O}, \mathcal{T}, \mathcal{R})$.

% \begin{itemize}[leftmargin=*, topsep=0pt, itemsep=0pt]
% \item \textbf{State Space ($\mathcal{S}$):} The latent, objective state of the environment (e.g., the complete underlying DOM structure, hidden backend states, or full retrieval databases).
% \item \textbf{Action Space ($\mathcal{A}$):} The set of admissible executable actions (e.g., \texttt{click(element)}, \texttt{type(text)}, \texttt{scroll()}, or submitting a search query).
% \item \textbf{Observation Space ($\mathcal{O}$):} At each discrete time step $t$, the environment emits a high-dimensional observation $o_t \in \mathcal{O}$ (e.g., an HTML accessibility tree or retrieved text paragraphs).
% \item \textbf{Transition $\mathcal{T}$ \& Reward $\mathcal{R}$:} The environment transitions according to $\mathcal{T}(s_{t+1}|s_t, a_t)$ and yields a sparse, terminal reward $R(\tau) \in \{0, 1\}$ upon task completion.
% \end{itemize}

We formulate the interaction between an autonomous LLM agent and a complex, long-horizon environment (e.g., a web browser or search engine) as a Partially Observable Markov Decision Process (POMDP), defined by the tuple $\mathcal{M} = (\mathcal{S}, \mathcal{A}, \mathcal{O}, \mathcal{T}, \mathcal{R})$. Here, $\mathcal{S}$ represents the latent, objective state of the environment (e.g., complete underlying DOM structures or full retrieval databases); $\mathcal{A}$ is the set of admissible executable actions (e.g., \texttt{click(element)}, \texttt{type(text)}, or search queries); and $\mathcal{O}$ denotes the high-dimensional observation space emitted at each discrete time step $t$ as $o_t \in \mathcal{O}$. Finally, the environment transitions according to $\mathcal{T}(s_{t+1}|s_t, a_t)$ and yields a sparse, terminal reward $R(\tau) \in \{0, 1\}$ upon task completion.

\textbf{The Standard Policy and the Context Bottleneck.}
\label{sec:preliminaries}
Given a user instruction $u$, a standard autoregressive LLM agent attempts to learn a policy $\pi_\theta(a_t | h_t)$, conditioned on the complete, uncompressed interaction history:
$
    h_t = \{u, o_0, a_0, o_1, a_1, \dots, o_t\}
$
The agent aims to maximize the expected return $J(\theta) = \mathbb{E}_{\tau \sim \pi_\theta} [R(\tau)]$. However, in real-world tasks, the observation $o_t$ is characterized by extreme informational entropy; it contains sparse, task-critical reasoning anchors $k_t$ buried within massive environmental noise $z_t$, where $|z_t| \gg |k_t|$. For example, in web navigation, a raw Document Object Model (DOM) tree ($o_t$) typically consists of over 90\% structural noise ($z_t$)—such as formatting tags, advertisements, and redundant navigational links—obscuring crucial, localized data points ($k_t$) like a specific product price or a hidden spatial coordinate.

As the temporal horizon $t$ expands, the sequence length of $h_t$ grows linearly. This raw concatenation triggers the \textbf{Context Bottleneck}: it inevitably distributes the model's probability mass across the accumulating noise $\sum z_t$, causing \textit{Attention Dilution} (the "Lost-in-the-Middle" phenomenon) and inducing \textit{Cascading Failures} as early noise compounds into hallucinated actions at step $t+1$.

\textbf{The Optimization Conflict of Monolithic Active Memory.}
To mitigate this unbounded context growth, recent RL-based architectures expand the agent's action space to include internal memory management operations: $\mathcal{A}' = \mathcal{A} \cup \mathcal{A}_{\text{mem}}$. The monolithic policy is reformulated to dynamically update a compressed state alongside executing environment actions: $\pi_\theta(a_t, m_t | \mathcal{M}_t, o_t)$. 

While this actively reduces sequence length, it introduces a severe \textit{optimization conflict}. Memory curation (which requires aggressive information entropy reduction) and task execution (which requires preserving deep logical dependencies) represent orthogonal cognitive objectives. Forcing a single parameter set $\theta$ to simultaneously optimize for both induces gradient interference, frequently diluting the reasoning trace and resulting in "lazy agent" behaviors. Furthermore, applying on-policy RL to align these dual objectives within massive, closed-source frontier models is computationally prohibitive.

\textbf{Objective of this Work: Symbiotic Decoupling via RL.}
To resolve this optimization conflict, our objective is to map raw history into a heavily condensed, high-fidelity memory state ($|\mathcal{M}_t| \ll |h_t|$) without burdening the primary reasoning engine. We achieve this by decoupling the monolithic policy $\pi_\theta(a_t, m_t)$ into two distinct, specialized entities: an \textit{active} context policy $\pi_{\text{ctx}}$ (ContextCurator) to handle $\mathcal{A}_{\text{mem}}$, and a \textit{frozen} execution policy $\pi_{\text{exec}}$ (TaskExecutor) to handle $\mathcal{A}$. 

The core technical challenge lies in cross-model alignment. We resolve this by absorbing the frozen TaskExecutor into the environment's transition dynamics, effectively reducing the symbiotic setup into a tractable single-agent RL problem. We then employ a Multi-Turn Group Relative Policy Optimization (MT-GRPO) pipeline to train a lightweight, open-source ContextCurator. By optimizing against distal rewards generated by this augmented environment, the Curator learns to perform targeted information entropy reduction perfectly aligned with the reasoning prerequisites of the black-box TaskExecutor.

\subsection{Symbiotic Cognitive Architecture}\label{sec:symbiotic_architecture}

We achieve this objective by decoupling the cognitive architecture into two specialized LLM components, ensuring that reasoning is not obstructed by low-level information overload. At any time step $t$, the symbiotic interaction unfolds in two stages:

\textbf{1. Active Context Curation ($\pi_{\text{ctx}}$).}
At each step $t$, the \textbf{ContextCurator} takes as input the current working memory $\mathcal{M}_t$, the latest observation $o_t$, and the
previous action $a_{t-1}$, and produces an updated memory state. Unlike prior active memory frameworks, such as Mem1~\citep{zhou2025mem1} and
MemoryR1~\citep{yan2025memory}, which formulate memory management as selecting from a small set of discrete operations (e.g., \texttt{navigate},
\texttt{scroll up}, and \texttt{rollback}), we instead model context curation as direct generation of the next memory state. Concretely, $
\pi_{\text{ctx}}$ autoregressively rewrites the current workspace into a refined textual memory $m_t$, which serves as the working memory for the
next step:
$
\mathcal{M}_{t+1} = m_t \sim \pi_{\text{ctx}}(\cdot \mid \mathcal{M}_t, o_t, a_{t-1}).
$
This formulation allows the curator to compress redundant interactions, remove structural noise, and preserve information that is most
relevant for subsequent reasoning and decision making.

\textbf{2. Context-Conditioned Execution ($\pi_{\text{exec}}$).}
Given the curated memory $m_t$ and the current observation $o_t$, the \textbf{TaskExecutor} generates the task action $a_t$. We instantiate the
executor with a strong frozen foundation model (e.g., GPT-4o or Gemini-3.0-flash), which conditions on both the curated memory and the immediate
environment signal to produce the next action:
$
a_t \sim \pi_{\text{exec}}(\cdot \mid m_t, o_t).
$
The quality of the generated action therefore depends critically on the fidelity and sufficiency of the curator's output.

\subsection{Policy Optimization via Multi-Turn GRPO}\label{sec:optimization}

While system prompts can define the functional role of the ContextCurator, they cannot inherently teach a small LLM \textit{what} information is causally salient for a given task. Relying solely on zero-shot prompting often leads to hallucinated pruning, where the model arbitrarily discards crucial reasoning anchors or retains irrelevant noise. To address this, we formulate context curation as a sequential decision-making problem and train the ContextCurator $\pi_{\phi}$ using an on-policy reinforcement learning approach based on Multi-Turn Group Relative Policy Optimization (MT-GRPO).

\textbf{Reward Definition and The Causal Dependency Chain.} 
Let $\tau = \{(o_t, m_t, a_t)\}_{t=0}^L$ be a sampled trajectory of length $L$. The environment provides a sparse binary reward $R(\tau) \in \{0, 1\}$ upon task completion. This reward is determined by task-specific outcome metrics, such as URL matches, string matches, or specific webpage state assertions. Since the TaskExecutor is strictly frozen, its policy $\pi_{\text{exec}}$ can be absorbed into the environment's transition dynamics. Consequently, any variance in the terminal reward $R(\tau)$ is entirely attributable to the quality of the curated memory $m_t$. Given local context $c_t = (\mathcal{M}_t, o_t, a_{t-1})$, the causal dependency chain is defined as:
\begin{equation}
    \underbrace{m_t \sim \pi_{\phi}(\cdot | c_t)}_{\text{Curator Decision}} \longrightarrow \underbrace{a_t \sim \pi_{\text{exec}}(\cdot | m_t, o_t)}_{\text{Executor Action}} \longrightarrow \underbrace{R(\tau)}_{\text{Distal Outcome}}
\end{equation}
\textbf{Trajectory Sampling and Advantage Estimation.}
For a given instruction $u$, we sample a group of $G$ independent trajectories $\{\tau_1, \tau_2, \dots, \tau_G\}$ generated by the symbiotic system. To reduce gradient variance over long horizons, we compute the group relative advantage $A_i$ for each trajectory $\tau_i$ by normalizing its reward against the group statistics:
\begin{equation}
\textstyle
    A_i = \frac{R(\tau_i) - \text{mean}(\{R(\tau_1), \dots, R(\tau_G)\})}{\text{std}(\{R(\tau_1), \dots, R(\tau_G)\}) + \epsilon}
\end{equation}
where $\epsilon$ is a small constant for numerical stability.

\textbf{Optimization Process and Policy Update.}
The ContextCurator's policy $\pi_{\phi}$ is updated to maximize this expected advantage. During the backward pass, gradients are applied \textit{only} to the tokens generated by the ContextCurator ($m_t$), dramatically reducing memory overhead compared to monolithic training. The multi-turn GRPO objective $\mathcal{J}_{\text{GRPO}}(\phi)$ is defined as the expectation over the sampled trajectories:
\begin{equation}
\label{eq:grpo_loss}
\textstyle
    \mathcal{J}_{\text{GRPO}}(\phi) = \mathbb{E}_{\tau \sim \pi_{\phi_{\text{old}}}} \left[ \frac{1}{L} \sum_{t=1}^{L} \left( \mathcal{O}_t^{\text{clip}}(\phi) - \beta \mathbb{D}_{\text{KL}} \big( \pi_\phi || \pi_{\text{ref}} \big)_t \right) \right]
\end{equation}
where the per-step clipped surrogate $\mathcal{O}_t^{\text{clip}}(\phi)$ restricts excessive policy updates:
\begin{equation}
\label{eq:clipped_surrogate}
    \mathcal{O}_t^{\text{clip}}(\phi) = \min \left( \rho_t(\phi) A_i, \text{clip}(\rho_t(\phi), 1-\epsilon, 1+\epsilon) A_i \right)
\end{equation}
Here, $\rho_t(\phi) = \frac{\pi_\phi(m_t | c_t)}{\pi_{\phi_{\text{old}}}(m_t | c_t)}$ is the importance sampling ratio of the Curator's actions, and $\beta$ controls the KL-divergence penalty against the reference model $\pi_{\text{ref}}$. 

Through this optimization process, the MT-GRPO objective forces the ContextCurator to perform active information entropy reduction. If $\pi_{\phi}$ retains visual noise or drops a reasoning anchor, the TaskExecutor fails, yielding $R(\tau)=0$. By optimizing against this distal reward, the Curator dynamically learns to adapt its context density, bridging the gap between raw environmental chaos and the strict reasoning prerequisites of the TaskExecutor.

\begin{table*}[t]
    \centering
    \small
    \setlength{\tabcolsep}{3.2pt}
    \caption[Experimental results of ActiveContext on WebArena.]{Experimental results on the WebArena benchmark. Success rate (SR $\uparrow$) and context token consumption (Tokens (K) $\downarrow$)\footnotemark~are reported across 5 domain subsets. To prevent extreme context bloat, the \textit{Full Context} baseline replaces historical webpage observations with placeholders. The bottom section extensively ablates different context curation methods using \texttt{gpt-4o-mini} as the fixed TaskExecutor. \textbf{The final average (Overall) is weighted by the number of instances in each subset.}}
    \label{tab:webarena}
    \resizebox{\textwidth}{!}{
    \begin{tabular}{l*{12}{c}}
        \toprule
        \multirow{2}{*}{\textbf{Methods}} & \multicolumn{2}{c}{\textbf{Shopping}} & \multicolumn{2}{c}{\textbf{Admin}} & \multicolumn{2}{c}{\textbf{Gitlab}} & \multicolumn{2}{c}{\textbf{Reddit}} & \multicolumn{2}{c}{\textbf{Map}} & \multicolumn{2}{c}{\textbf{Overall}} \\
        \cmidrule(lr){2-3} \cmidrule(lr){4-5} \cmidrule(lr){6-7} \cmidrule(lr){8-9} \cmidrule(lr){10-11} \cmidrule(lr){12-13}
        & \textit{SR} & \textit{Tokens} & \textit{SR} & \textit{Tokens} & \textit{SR} & \textit{Tokens} & \textit{SR} & \textit{Tokens} & \textit{SR} & \textit{Tokens} & \textit{SR} & \textit{Tokens} \\
        
        \midrule
        
        \multicolumn{13}{c}{\cellcolor{gray!15}\textbf{TaskExecutor: Gemini-3.0-flash}} \\
        No Memory& 13.0 & 50.7 & 22.2 & 57.2 & 36.7 & 75.1 & 26.1 & 42.8 & 6.7 & \underline{27.2} & 20.0 & 51.2 \\
        Full Context & \textbf{28.3} & 42.2 & 33.3 & 58.6 & \textbf{66.7} & 58.3 & 52.2 & \underline{35.5} & 10.0 & 40.2 & 36.4 & 47.4 \\
        Synapse & 19.6 & 44.6 & \textbf{41.7} & 71.5 & 43.3 & 70.0 & \textbf{56.5} & 44.4 & 26.7 & 36.6 & 35.2 & 53.6\\
        Training-Free GRPO & 17.4 & 55.4 & 36.1 & 70.6 & 36.7 & 76.9 & \textbf{56.5} & 47.8 & 30.0 & 50.1 & 32.7 & 60.6\\
        \textbf{ActiveContext (Ours)} & 26.1 & \underline{39.0} & 41.7 & \underline{50.5} & \textbf{66.7} & \underline{53.8} & 47.8 & 43.1 & \textbf{33.3} & 30.9 & \textbf{41.2} & \underline{43.3} \\
        
        \midrule
    
        \multicolumn{13}{c}{\cellcolor{gray!15}\textbf{TaskExecutor: Gemini-2.5-flash}} \\
        No Memory&   10.9 & 45.2 & 11.1 & 75.0 & 36.7 & 71.1 & 26.1 & 37.7 & 6.7 & 36.4 & 17.0 & 53.8 \\
        Full Context & 21.7 & 31.9 & 33.3 & 52.7 & 50.0 & 66.5 & 39.1 & \underline{32.6} & 13.3 & 30.4 & 30.3 & 42.6 \\
        Synapse & 13.0 & \underline{27.2} & \textbf{41.7} & 54.0 & 40.0 & 70.2 & 47.8 & 33.3 & 13.3 & 33.5 & 29.1 & 42.9\\
        Training-Free GRPO & 21.7 & 51.7 & 36.1 & 62.3 & 33.3 & 80.6 & \textbf{52.2} & 43.7 & 3.3 & 64.4 & 27.9 & 60.5 \\
        \textbf{ActiveContext (Ours)} & \textbf{23.9} & 28.0 & 33.3 & \underline{51.6} & \textbf{50.0} & \underline{53.1} & 47.8 & 34.3 & \textbf{13.3} & \underline{28} & \textbf{32.1} & \underline{38.6} \\
    
        \midrule
    
        \multicolumn{13}{c}{\cellcolor{gray!15}\textbf{TaskExecutor: gpt-4o}} \\
        No Memory & 13.0 & 49.5 & \textbf{33.3} & 63.9 & 33.3 & 73.9 & 47.8 & 35.6 & 6.7 & 35.8 & 24.8 & 52.6 \\
        Full Context &13.0 & 38.9 & 25 & \underline{46.4} & 53.3 & \underline{54.1} & 43.5 & \underline{27.5} & 13.3 & 37.5 & 27.3 & 41.5 \\
        Synapse & 17.4 & \underline{35.2} & 16.7 & 65.6 & 23.3 & 67.0 & 47.8 & 35.7 & 20.0 & 26.9 & 23.0 & 46.2\\
        Training-Free GRPO & 10.9 & 50.8 & 27.8 & 53.8 & 33.3 & 64.8 & 43.5 & 41.4 & \textbf{23.3} & 31.3 & 25.5 & 49.1 \\
        \textbf{ActiveContext (Ours)} &\textbf{19.6} & 40.6 & 22.2 & \underline{46.4} & \textbf{56.7} & 54.7 & \textbf{52.2} & 33.8 & 13.3 & \underline{25.9} & \textbf{30.3} & \underline{40.8} \\
    
        \midrule

        \multicolumn{13}{c}{\cellcolor{gray!15}\textbf{TaskExecutor: gpt-4o-mini} (with Extended Ablations)} \\
        \multicolumn{13}{c}{\textbf{\textit{Baselines \& Retrieval-Based Memory}}} \\
        No Memory& 8.7 & 80.3 & 8.3 & 100.3 & \textbf{26.7} & 87.4 & 8.7 & 62.2 & 3.3 & 59.4 & 10.9 & 79.6 \\
        Full Context & 8.7 & 46.3 & \textbf{16.7} & \underline{51.0} & 23.3 & \underline{68.2} & 13.0 & \underline{43.0} & 3.3 & 36.9 & 12.7 & \underline{49.1} \\
        Synapse & \textbf{10.9} & \underline{40.5} & 11.1 & 65.2 & 16.7 & 73.2 & \textbf{39.1} & 43.6 & 16.7 & \underline{25.6} & 17.0 & 49.6\\
        AWM & 8.7 & 51.3 & 11.1 & 68.7 & 20.0 & 75.6 & 34.8 & 44.2 & 16.7 & 34.7  &16.4 & 55.5\\
        Mem0 & 4.3 & 44.7 & \textbf{16.7} & 71.2 & 20.0 & 76.8 & 26.1 & 50.3 & 16.7 & 33.6 & 15.2 & 55.1 \\
        Training-Free GRPO & \textbf{10.9} & 60.0 & 13.9 & 71.1 & \textbf{26.7} & 78.7 & \textbf{39.1} & 47.5 & \textbf{20.0} & 42.1 & \textbf{20.0} & 60.8  \\
        \multicolumn{13}{c}{\textbf{\textit{Prompting-Based Memory Management (Zero-Shot Curators)}}} \\
        Gemini-3.0-flash & \textbf{17.4} & 37.8 & \textbf{25.0} & 54.6 & \textbf{40.0} & \underline{48.1} & 30.4 & 26.1 & 10.0 & 32.0 & \textbf{23.6} & 40.7 \\
        Gemini-2.5-flash & 13.0 & \underline{30.2} & 22.2 & \underline{38.3} & 33.3 & 48.6 & 30.4 & \underline{25.3} & 10.0 & \underline{18.7} & 20.6 & \underline{32.5} \\
        gpt-4o & 15.2 & 42.6 & 22.2 & 48.8 & 33.3 & 56.1 & \textbf{39.1} & 30.5 & 3.3 & 28.8 & 21.2 & 42.2 \\
        gpt-4o-mini & 10.9 & 43.7 & 11.1 & 51.9 & 26.7 & 58.3 & 21.7 & 33.5 & 6.7 & 31.2 & 14.6 & 44.4 \\
        Qwen2.5-7B-Instruct & 13.0 & 42.7 & 8.3 & 59.0 & 33.3 & 72.8 & 13.0 & 47.9 & 3.3 & 38.2 & 13.9 & 51.6 \\
        \multicolumn{13}{c}{\textbf{\textit{Active Memory Management (RL-Trained Curator)}}} \\
        \textbf{ContextCurator (Ours)} & 13.0 & \underline{30.2} & 22.2 & \underline{38.3} & 33.3 & \underline{48.6} & 39.1 & \underline{25.3} & 10.0 & \underline{18.7} & 21.8 & \underline{32.5} \\
        \bottomrule
    \end{tabular}
    }
    \vspace{-10pt}
\end{table*}
\footnotetext{Token counts are aggregated across the entire interaction trajectory. Because different methods require a varying number of steps to complete a task, the \textit{Full Context} baseline may occasionally accumulate fewer total tokens than \textit{No Memory}, as the latter typically requires longer exploration to succeed.}

\vspace{-5pt}
\section{Experiments}\label{sec:experiments}
\vspace{-3pt}

\subsection{Experimental Setup}
\textbf{Benchmarks.} We evaluate ActiveContext on two challenging long-horizon benchmarks:
(1) \textit{WebArena}~\citep{zhou2024webarena}: A realistic execution environment simulating four domains (Shopping, Forums, Gitlab, CMS), evaluating the agent's ability to navigate raw DOM trees over extended trajectories characterized by high structural noise.
(2) \textit{DeepSearch}~\citep{Search-R1}: A complex RAG-based environment requiring multi-turn retrieval and reasoning. It specifically tests the agent's ability to synthesize answers from noisy search engine results across datasets like HotpotQA and 2WikiMultiHop. We report the Success Rate (SR) and the total context token consumption across both environments. The exact formulation for calculating context tokens is detailed in Appendix~\ref{sec:context_calculation}.

\textbf{Baselines \& Models.} We compare ActiveContext against two standard baselines: \textit{No Memory}, where the policy conditions solely on the immediate observation, and \textit{Full Context}, which concatenates the entire interaction history. To mitigate extreme context bloat in the web environment, the \textit{Full Context} baseline specifically replaces verbose historical observations with placeholders. Additionally, we evaluate against advanced memory architectures, including the trajectory-based Synapse~\citep{zheng2024synapse}, the workflow-based AWM~\citep{wang2025agent}, Mem0~\citep{chhikara2025mem0} and Training-Free GRPO~\citep{25tfgrpo}. For our on-policy RL training, we initialize our specialized policy model (ContextCurator) with \texttt{Qwen-2.5-7B-Instruct}~\citep{qwen_2_5}. To assess cross-model generalization, we evaluate the TaskExecutor across a spectrum of proprietary frontier models: \texttt{Gemini-3.0-flash}, \texttt{Gemini-2.5-flash}~\citep{comanici2025gemini}, \texttt{GPT-4o}, and \texttt{GPT-4o-mini}~\citep{gpt4o}. Crucially, aligning with our mathematical formulation, these proprietary models remain strictly frozen and are absorbed into the environment's transition dynamics during training, necessitating maximal stability.

\textbf{Training Details.} Our policy model is trained on a single node equipped with 8 NVIDIA A100 GPUs (80GB), using full-parameter fine-tuning. For the MT-GRPO training, we utilize a constant learning rate of $1e-6$ with a batch size of 8. The rollout number $G$ is set to 4 for WebArena-Lite and 8 for DeepSearch. In both tasks, the number of interaction rounds with environment is capped at 15. The KL divergence regularization coefficient $\beta$ and the clip ratio $\epsilon$ are set to 0.001 and 0.2, respectively. The maximum context length and the number of new tokens are configured to 16,000 and 512. For efficient LLM rollouts during RL, we leverage vLLM with a tensor parallel size of 1 and a GPU memory utilization ratio of 0.6. Rollout sampling is performed with both the temperature and top-$p$ set to 1.0.

\vspace{-8pt}
\subsection{Results on WebArena Benchmark}\label{sec:exp_web}
\vspace{-4pt}

Table~\ref{tab:webarena} presents the comparative evaluation on WebArena. The results empirically validate our primary hypothesis: \textit{active context curation elevates the reasoning upper-bound while drastically reducing computational costs}.

\textbf{Context Memory Quality Matters.}
Standard approaches face a severe dilemma: \textit{No Memory} lacks historical environment feedback, while \textit{Full Context} induces Attention Dilution due to context overload. ActiveContext breaks this trade-off, establishing a new Pareto frontier. With \textit{Gemini-3.0-flash}, our framework achieves a Success Rate (SR) of \textbf{41.2\%} (vs. 36.4\% with full context) while simultaneously reducing token consumption by \textbf{8.8\%} (43.3k vs. 47.4k). This confirms that the bottleneck in web agents is not the \textit{quantity} of information, but the \textit{Signal-to-Noise Ratio (SNR)}. By actively pruning redundant DOM elements (e.g., ads, sidebars), ActiveContext prevents cascading failures, allowing the backbone's attention mechanism to focus entirely on critical decision pivots.

\textbf{Preserving Reasoning Anchors.}
In the \textbf{Map} domain, which requires precise retention of spatial coordinates and user arguments over long horizons, \textit{No Memory} often fails near-completely (e.g., 6.7\% SR for \textit{Gemini-3.0-flash}). In contrast, ActiveContext achieves up to \textbf{33.3\%} SR. This substantial gap confirms that ActiveContext successfully captures implicit "Reasoning Anchors"—data points like spatial coordinates that appear locally irrelevant but are globally causal—which heuristic summarization or full-context approaches routinely discard or bury.

\textbf{ContextCurator Efficacy Across Model Scales.}
The performance benefits of ActiveContext are disproportionately higher for smaller backbone models. For \textit{GPT-4o-mini}, the framework improves the SR from 12.7\% to \textbf{21.8\%} (a +72\% relative gain), whereas \textit{GPT-4o} achieves an 11\% relative gain. This disparity suggests that while frontier models possess some inherent robustness to structural noise, smaller models rely heavily on the ContextCurator to distill a high-fidelity workspace for successful task execution.

\textbf{Impact of MT-GRPO Optimization.}
Table~\ref{tab:webarena} isolates the contribution of our RL formulation by fixing \texttt{gpt-4o-mini} as the TaskExecutor and evaluating distinct memory management strategies. We compare retrieval-based memories (e.g., Synapse, Training-Free GRPO) against prompted active memory management (where proprietary LLMs act as the ContextCurator using our standard prompt), and finally against our RL-optimized ContextCurator. Comparing our RL-finetuned \texttt{ContextCurator} against its base model, \texttt{Qwen2.5-7B-Instruct}, we observe a significant leap in SR (13.9\% $\to$ \textbf{21.8\%}) alongside a massive reduction in context token consumption (54.4k $\to$ 32.5k). Crucially, our \texttt{ContextCurator} substantially outperforms the proprietary \texttt{gpt-4o-mini} context agent (21.8\% vs 14.6\%) while using fewer tokens (32.5k vs 44.4k). Furthermore, our lightweight 7B model achieves competitive parity with the massively larger \texttt{gpt-4o} acting as a curator (21.2\% SR). This validates that specialization via distal RL rewards enables open-source small models to act as highly effective cognitive filters, replacing expensive proprietary models for the context management role.

\begin{table*}[t]
    \centering
    \setlength{\tabcolsep}{2.0pt} 
    \small
    \setlength{\belowcaptionskip}{0.0pt}
    \caption{Experimental results of \textit{ActiveContext} on the DeepSearch benchmark. Success rate (SR $\uparrow$) and context token consumption (Tokens (K) $\downarrow$) are reported across 7 domain subsets for 4 different backbone LLMs. The bottom section extensively ablates different context curation methods using \texttt{gpt-4o-mini} as the fixed TaskExecutor.}
    \label{tab:search}
    \resizebox{\textwidth}{!}{
    \begin{tabular}{l*{16}{c}}
        \toprule
        \multirow{2}{*}{\textbf{Methods}} & \multicolumn{2}{c}{\textbf{NQ}} & \multicolumn{2}{c}{\textbf{TriviaQA}} & \multicolumn{2}{c}{\textbf{PopQA}} & \multicolumn{2}{c}{\textbf{HotpotQA}} & \multicolumn{2}{c}{\textbf{2wiki}} & \multicolumn{2}{c}{\textbf{Musique}} & \multicolumn{2}{c}{\textbf{Bamboogle}} & \multicolumn{2}{c}{\textbf{Avg.}} \\
        \cmidrule(lr){2-3} \cmidrule(lr){4-5} \cmidrule(lr){6-7} \cmidrule(lr){8-9} \cmidrule(lr){10-11} \cmidrule(lr){12-13} \cmidrule(lr){14-15} \cmidrule(lr){16-17}
         & \textit{SR} & \textit{Tokens} & \textit{SR} & \textit{Tokens} & \textit{SR} & \textit{Tokens} & \textit{SR} & \textit{Tokens} & \textit{SR} & \textit{Tokens} & \textit{SR} & \textit{Tokens} & \textit{SR} & \textit{Tokens} & \textit{SR} & \textit{Tokens} \\
        \midrule
        \multicolumn{17}{c}{\cellcolor{gray!15}\textbf{TaskExecutor: Gemini-3.0-flash}} \\
        Full Context & 30.0 & 62.4 & 76.0 & 20.4 & 62.0 & 36.4 & 52.0 & 53.5 & 62.0 & 38.9 & 33.0 & 83.6 & 62.0 & 31.6 & 53.9 & 46.7 \\
        Synapse & \textbf{86.0} & 53.6 & 68.0 & 51.2 & 54.0 & 100.5 & 32.0 & 149.2 & 42.0 & 83.0 & \textbf{44.0} & 83.0 & \textbf{72.0} & 44.6 & 56.9 & 80.7 \\
        Training-Free GRPO & 22.0 & 78.2 & 76.0 & 16.7 & 44.0 & 47.2 & 46.0 & 60.5 & \textbf{68.0} & 39.1 & 27.0 & 82.7 & 60.0 & 33.4 & 49.0 & 51.1 \\
        \textbf{ActiveContext (Ours)} & 32.0 & \underline{6.8} & \textbf{78.0} & \underline{3.2} & \textbf{64.0} & \underline{7.4} & \textbf{56.0} & \underline{7.4} & \textbf{68.0} & \underline{7.5} & 36.0 & \underline{8.5} & 66.0 & \underline{5.6} & \textbf{57.1} & \underline{6.6} \\
        
        \midrule
        
        \multicolumn{17}{c}{\cellcolor{gray!15}\textbf{TaskExecutor: Gemini-2.5-flash}} \\
        Full Context & 22.0 & 38.2 & 58.0 & 20.6 & 32.0 & 35.1 & \textbf{32.0} & 35.5 & 28.0 & 40.8 & 16.0 & 42.0 & 46.0 & 28.8 & 33.4 & 34.4 \\
        Synapse & \textbf{42.0} & 81.4 & 35.0 & 86.1 & \textbf{48.0} & 76.4 & 26.0 & 135.3 & 32.0 & 92.4 & \textbf{40.0} & 77.8 & \textbf{62.0} & 52.0 & 40.7 & 85.9 \\
        Training-Free GRPO & 10.0 & 81.2 & 60.0 & 38.4 & 26.0 & 75.9 & 28.0 & 76.5 & 24.0 & 78.3 & 17.0 & 119.0 & 42.0 & 63.5 & 29.6 & 76.1 \\
        \textbf{ActiveContext (Ours)} & 27.9 & \underline{8.1} & \textbf{66.0} & \underline{3.7} & 40.9 & \underline{7.3} & 31.8 & \underline{8.4} & \textbf{36.4} & \underline{9.6} & 25.6 & \underline{8.3} & 61.9 & \underline{5.6} & \textbf{41.5} & \underline{7.3} \\
        \midrule
        \multicolumn{17}{c}{\cellcolor{gray!15}\textbf{TaskExecutor: gpt-4o}} \\
        Full Context & 22.0 & 40.8 & 58.0 & 20.7 & 32.0 & 35.1 & 32.0 & 35.5 & \textbf{28.0} & 38.2 & 16.0 & 42.0 & 46.0 & 28.8 & 33.4 & 34.4 \\
        Synapse & \textbf{48.0} & 38.9 & 36.0 & 42.0 & \textbf{40.0} & 47.8 & 18.0 & 52.1 & \textbf{28.0} & 45.9 & \textbf{46.0} & 38.9 & \textbf{58.0} & 27.4 & \textbf{39.1} & 41.9 \\
        Training-Free GRPO & 26.0 & 35.3 & 58.0 & 19.6 & 36.0 & 31.0 & 34.0 & 35.5 & \textbf{28.0} & 38.9 & 15.0 & 42.8 & 34.0 & 31.9 & 33.0 & 33.6\\
        \textbf{ActiveContext (Ours)} & 28.0 & \underline{7.2} & \textbf{66.0} & \underline{3.6} & 26.0 & \underline{7.7} & \textbf{38.0} & \underline{7.4} & 26.0 & \underline{9.3} & 15.0 & \underline{9.1} & 44.0 & \underline{6.4} & 34.7 & \underline{7.2} \\
        \midrule
        \multicolumn{17}{c}{\cellcolor{gray!15}\textbf{TaskExecutor: gpt-4o-mini}(with Extended Ablations)}  \\
        \multicolumn{17}{c}{\textbf{\textit{Baselines \& Retrieval-Based Memory}}} \\
        Full Context & 20.0 & \underline{35.5} & 50.0 & 19.7 & 34.0 & \underline{33.8} & {30.0} & 37.2 & {40.0} & \underline{33.8} & 15.0 & 42.1 & 40.0 & \underline{30.0} & 32.7 & \underline{33.2} \\
        Synapse & \textbf{52.0} & 38.6 & 37.0 & 42.7 & \textbf{38.0} & 50.5 & 20.0 & 55.9 & 38.0 & 45.0 & \textbf{42.0} & \underline{41.8} & \textbf{60.0} & 30.9 & 41.0 & 43.6 \\
        AWM & 24.0 & 39.1 & 56.0 & \underline{16.4} & 34.0 & 34.8 & 25.0 & \underline{36.2} & 39.0 & 34.2 & 18.0 & 43.2 & 38.0 & 32.1 & 33.4 &  33.7 \\
        Mem0 & 23.0 & 46.7 & \textbf{62.0} & 27.6 & 32.0 & 47.9 & \textbf{31.0} & 52.1 & \textbf{42.0} & 46.7 & 15.0 & 59.7 & 40.0 & 45.4 & 34.7 & 46.6 \\
        Training-Free GRPO & 20.0 & 40.1 & 64.0 & 18.3 & 32.0 & 36.2 & 30.0 & 39.1 & 40.0 & 36.1 & 16.0 & 45.4 & 34.0 & 35.0 & 33.7 & 35.7 \\
        \multicolumn{17}{c}{\cellcolor{gray!15}\textbf{\textit{Prompting-based Memory Management(Zero-Shot Curators)}}} \\
        Gemini-3.0-flash & \textbf{34.0} & 8.0 & \textbf{66.0} & \underline{4.4} & \textbf{40.0} & 8.0 & \textbf{38.0} & 8.7 & \textbf{42.0} & 9.5 & \textbf{19.0} & 9.6 & \textbf{56.0} & 6.3 & \textbf{42.1} & 7.8 \\
        Gemini-2.5-flash & 28.0 & 7.9 & 64.0 & 4.6 & 30.0 & 8.2 & 32.0 & 8.3 & 34.0 & 9.1 & 17.0 & 9.7 & 52.0 & \underline{6.1} & 36.7 & 7.7 \\
        gpt-4o & 30.0 & 11.2 & 58.0 & 6.9 & 34.0 & 12.0 & 30.0 & 11.4 & 32.0 & 11.2 & 17.0 & 13.4 & 44.0 & 9.2 & 35.0 & 10.8 \\
        gpt-4o-mini &28.0 & 11.2 & 56.0 & 7.3 & 32.0 & 10.8 & 28.0 & 10.9 & 30.0 & 10.0 & 17.0 & 12.6 & 40.0 & 8.9 & 33.0 & 10.2 \\
        Qwen2.5-7B-Instruct &22.0 & 10.7 & 52.0 & 6.5 & 28.0 & 9.9 & 28.0 & 10.0 & 28.6 & 10.5 & 10.0 & 12.9 & 44.0 & 8.7 & 30.4 & 9.9 \\
        \multicolumn{17}{c}{\textbf{\textit{Active Memory Management (RL-Trained Curator)}}} \\
        \textbf{ContextCurator (Ours)} & 28.0 & \underline{7.0} & 62.0 & 5.0 & 38.0 & \underline{7.6} & 34.0 & \underline{7.3} & 38.0 & \underline{7.4} & 12.0 & \underline{9.3} & 32.0 & 7.3 & 34.9 & \underline{7.3} \\ 
      \bottomrule
    \end{tabular}
    }
    \vspace{-10pt}
\end{table*}

\vspace{-3pt}
\subsection{Results on DeepSearch Benchmark}\label{sec:exp_search}
\vspace{-2pt}

Table~\ref{tab:search} presents results on DeepSearch. Unlike WebArena's visual structural noise, DeepSearch introduces severe \textit{semantic noise}, where redundant RAG results degrade the model's attention mechanism and impair subsequent logic. 

\textbf{Information Entropy Reduction in Search.}
\textit{ActiveContext} consistently outperforms the \textit{Full Context} baselines. When utilizing \textit{Gemini-2.5-flash} as the executor, it achieves a \textbf{24\%} relative improvement in success rate (33.4\% $\to$ 41.5\%) while simultaneously reducing the average context length by a factor of \textbf{$\sim5\times$} (34.4k $\to$ 7.3k). By aggressively stripping away semantic noise, \textit{ActiveContext} helps the executor maintain a pristine reasoning trace. This confirms that for complex multi-turn search tasks, actively filtering irrelevant paragraphs to maximize the Signal-to-Noise Ratio (SNR) in working memory is imperative for preserving logical coherence. We note that the trajectory-based \textit{Synapse} method is also quite effective on these QA benchmarks; because many queries share underlying semantic structures, retrieving a successful historical trajectory as an exemplar can significantly boost performance. However, this comes at the cost of massive context bloat.

\textbf{Domain Analysis: Handling Multi-Hop Complexity.}
The impact of ActiveContext is most pronounced in multi-hop reasoning tasks like \textit{2WikiMultiHop} and \textit{HotpotQA}. As shown in Table~\ref{tab:search}, for Gemini-3.0-flash executor, ActiveContext improves \textit{2Wiki} performance from 62.0\% to \textbf{68.0\%}, and \textit{HotpotQA} from 52.0\% to \textbf{56.0\%}. In these domains, distractor paragraphs often mislead the model into incorrect intermediate reasoning steps. ActiveContext effectively filters these distractors, preserving only the golden evidence. In contrast, simpler single-hop tasks like \textit{NQ} (Natural Questions) show smaller gains (30.0\% $\to$ 32.0\%), suggesting that active curation is most critical when the reasoning chain is fragile and highly susceptible to noise accumulation.

\textbf{RL-Driven Cognitive Decoupling.}
Table~\ref{tab:search} provides strong evidence for our "Cognitive Decoupling" motivation in the search domain. The base \texttt{Qwen2.5-7B-Instruct} model (30.4\% SR) underperforms the \textit{Full Context} baseline (32.7\% SR), indicating that generic instructions are insufficient for small LLMs to effectively distinguish signal from semantic noise in complex retrieved documents. However, the MT-GRPO optimization improves this performance to \textbf{34.9\%}, proving that the RL objective successfully instills a robust, global context memory management capability. Remarkably, the specialized ContextCurator outperforms the generic \texttt{gpt-4o-mini} agent (33.0\%) and achieves effective parity with the much larger \texttt{gpt-4o} (35.0\%), all while consuming significantly fewer tokens (7.3k vs 10.8k).

\textbf{Case Study.} 
As detailed in Appendix~\ref{sec:case_study}, ActiveContext's performance on the WebArena benchmark practically demonstrates its ability to organically condense multi-turn interaction history into concise, actionable memory. This mechanism ensures the GPT-4o-mini model maintains focus on the distal task objective, avoids getting lost in verbose DOM observations, and efficiently arrives at the correct answer during complex multi-step navigation. 

Collectively, these results across the Web and Search domains demonstrate that ActiveContext is not merely a compression heuristic, but a fundamental \textit{reasoning enhancer}. By decoupling context management from execution and optimizing the former via reinforcement learning, we establish a highly scalable, efficient, and performant paradigm for autonomous long-horizon agents.

\vspace{-10pt}
\section{Conclusion}
\vspace{-8pt}

In this work, we address the critical "Context Bottleneck" that limits the performance of Large Language Model (LLM) agents in long-horizon environments. We introduce ActiveContext, a symbiotic framework that redefines context memory management as an active, sequential decision-making process, moving beyond the traditional static storage paradigm. By decoupling the ContextCurator (Context Architect) from the TaskExecutor (Reasoning Engine) and leveraging on-policy reinforcement learning, we transform memory into a highly dynamic and efficient control variable.

Our empirical evaluations across the WebArena and DeepSearch benchmarks demonstrate the efficacy of this paradigm. ActiveContext establishes a new Pareto frontier, significantly enhancing the reasoning upper-bound of state-of-the-art (SOTA) models while drastically reducing token consumption—achieving up to an 85\% reduction in retrieval-heavy environments (e.g., Gemini-3.0-flash on DeepSearch). Notably, a specialized, RL-finetuned 7B model can match the context management performance of proprietary frontier models like GPT-4o, underscoring that active context curation is a distinct cognitive capability separate from general reasoning.

We believe this "Specialized Symbiosis" framework provides a foundational blueprint for the next generation of sustainable, high-fidelity autonomous agents. It shifts the field's focus from merely brute-forcing larger context windows to intelligently managing an agent's cognitive span, ensuring agents can effectively handle complex, long-horizon tasks with superior efficiency and reasoning quality.

% \section*{Author Contributions}

% \section*{Acknowledgments}
% Use unnumbered first level headings for the acknowledgments. All

% \section*{Ethics Statement}
% Authors can add an optional ethics statement to the paper. 
% For papers that touch on ethical issues, this section will be evaluated as part of the review process. The ethics statement should come at the end of the paper. It does not count toward the page limit, but should not be more than 1 page. 

\bibliography{colm2026_conference}

@misc{o1,
  title={Learning to reason with LLMs},
  author={OpenAI},
  year={2024},
  url = {https://openai.com/index/learning-to-reason-with-llms/}
}

@misc{grpo,
      title={DeepSeekMath: Pushing the Limits of Mathematical Reasoning in Open Language Models}, 
      author={Zhihong Shao and Peiyi Wang and Qihao Zhu and Runxin Xu and Junxiao Song and Xiao Bi and Haowei Zhang and Mingchuan Zhang and Y. K. Li and Y. Wu and Daya Guo},
      year={2024},
      eprint={2402.03300},
      archivePrefix={arXiv},
      primaryClass={cs.CL},
      url={https://arxiv.org/abs/2402.03300}, 
}

@misc{ppo,
      title={Proximal Policy Optimization Algorithms}, 
      author={John Schulman and Filip Wolski and Prafulla Dhariwal and Alec Radford and Oleg Klimov},
      year={2017},
      eprint={1707.06347},
      archivePrefix={arXiv},
      primaryClass={cs.LG},
      url={https://arxiv.org/abs/1707.06347}, 
}

@inproceedings{yang2024sweagent,
    author = {John Yang and
Carlos E. Jimenez and
Alexander Wettig and
Kilian Lieret and
Shunyu Yao and
Karthik Narasimhan and
Ofir Press},
    bibsource = {dblp computer science bibliography, https://dblp.org},
    booktitle = {Advances in Neural Information Processing Systems 38: Annual Conference
on Neural Information Processing Systems 2024, NeurIPS 2024, Vancouver,
BC, Canada, December 10 - 15, 2024},
    editor = {Amir Globersons and
Lester Mackey and
Danielle Belgrave and
Angela Fan and
Ulrich Paquet and
Jakub M. Tomczak and
Cheng Zhang},
    title = {SWE-agent: Agent-Computer Interfaces Enable Automated Software Engineering},
    year = {2024}
}

@inproceedings{gur2024a,
    author = {Izzeddin Gur and
Hiroki Furuta and
Austin V. Huang and
Mustafa Safdari and
Yutaka Matsuo and
Douglas Eck and
Aleksandra Faust},
    bibsource = {dblp computer science bibliography, https://dblp.org},
    booktitle = {The Twelfth International Conference on Learning Representations,
{ICLR} 2024, Vienna, Austria, May 7-11, 2024},
    publisher = {OpenReview.net},
    title = {A Real-World WebAgent with Planning, Long Context Understanding, and
Program Synthesis},
    url = {https://openreview.net/forum?id=9JQtrumvg8},
    year = {2024}
}

@article{comanici2025gemini,
    author = {Comanici, Gheorghe and Bieber, Eric and Schaekermann, Mike and Pasupat, Ice and Sachdeva, Noveen and Dhillon, Inderjit and Blistein, Marcel and Ram, Ori and Zhang, Dan and Rosen, Evan and others},
    journal = {ArXiv preprint},
    title = {Gemini 2.5: Pushing the frontier with advanced reasoning, multimodality, long context, and next generation agentic capabilities},
    url = {https://arxiv.org/abs/2507.06261},
    volume = {abs/2507.06261},
    year = {2025}
}

@inproceedings{xie2024osworld,
    author = {Tianbao Xie and
Danyang Zhang and
Jixuan Chen and
Xiaochuan Li and
Siheng Zhao and
Ruisheng Cao and
Toh Jing Hua and
Zhoujun Cheng and
Dongchan Shin and
Fangyu Lei and
Yitao Liu and
Yiheng Xu and
Shuyan Zhou and
Silvio Savarese and
Caiming Xiong and
Victor Zhong and
Tao Yu},
    bibsource = {dblp computer science bibliography, https://dblp.org},
    booktitle = {Advances in Neural Information Processing Systems 38: Annual Conference
on Neural Information Processing Systems 2024, NeurIPS 2024, Vancouver,
BC, Canada, December 10 - 15, 2024},
    editor = {Amir Globersons and
Lester Mackey and
Danielle Belgrave and
Angela Fan and
Ulrich Paquet and
Jakub M. Tomczak and
Cheng Zhang},
    title = {OSWorld: Benchmarking Multimodal Agents for Open-Ended Tasks in Real
Computer Environments},
    year = {2024}
}

@article{liu2025advances,
    author = {Liu, Bang and Li, Xinfeng and Zhang, Jiayi and Wang, Jinlin and He, Tanjin and Hong, Sirui and Liu, Hongzhang and Zhang, Shaokun and Song, Kaitao and Zhu, Kunlun and others},
    journal = {ArXiv preprint},
    title = {Advances and challenges in foundation agents: From brain-inspired intelligence to evolutionary, collaborative, and safe systems},
    url = {https://arxiv.org/abs/2504.01990},
    volume = {abs/2504.01990},
    year = {2025}
}

@article{zhang2024survey,
    author = {Zhang, Zeyu and Dai, Quanyu and Bo, Xiaohe and Ma, Chen and Li, Rui and Chen, Xu and Zhu, Jieming and Dong, Zhenhua and Wen, Ji-Rong},
    journal = {ACM Transactions on Information Systems},
    publisher = {ACM New York, NY},
    title = {A survey on the memory mechanism of large language model based agents},
    year = {2024}
}

@inproceedings{zhong2024memorybank,
    author = {Wanjun Zhong and
Lianghong Guo and
Qiqi Gao and
He Ye and
Yanlin Wang},
    bibsource = {dblp computer science bibliography, https://dblp.org},
    booktitle = {Thirty-Eighth {AAAI} Conference on Artificial Intelligence, {AAAI}
2024, Thirty-Sixth Conference on Innovative Applications of Artificial
Intelligence, {IAAI} 2024, Fourteenth Symposium on Educational Advances
in Artificial Intelligence, {EAAI} 2014, February 20-27, 2024, Vancouver,
Canada},
    doi = {10.1609/AAAI.V38I17.29946},
    editor = {Michael J. Wooldridge and
Jennifer G. Dy and
Sriraam Natarajan},
    pages = {19724--19731},
    publisher = {{AAAI} Press},
    title = {MemoryBank: Enhancing Large Language Models with Long-Term Memory},
    url = {https://doi.org/10.1609/aaai.v38i17.29946},
    year = {2024}
}

@article{zhou2025mem1,
    author = {Zhou, Zijian and Qu, Ao and Wu, Zhaoxuan and Kim, Sunghwan and Prakash, Alok and Rus, Daniela and Zhao, Jinhua and Low, Bryan Kian Hsiang and Liang, Paul Pu},
    journal = {ArXiv preprint},
    title = {MEM1: Learning to Synergize Memory and Reasoning for Efficient Long-Horizon Agents},
    url = {https://arxiv.org/abs/2506.15841},
    volume = {abs/2506.15841},
    year = {2025}
}

@article{yu2025memagent,
    author = {Yu, Hongli and Chen, Tinghong and Feng, Jiangtao and Chen, Jiangjie and Dai, Weinan and Yu, Qiying and Zhang, Ya-Qin and Ma, Wei-Ying and Liu, Jingjing and Wang, Mingxuan and others},
    journal = {ArXiv preprint},
    title = {MemAgent: Reshaping Long-Context LLM with Multi-Conv RL-based Memory Agent},
    url = {https://arxiv.org/abs/2507.02259},
    volume = {abs/2507.02259},
    year = {2025}
}

@misc{packer2023memgpt,
      title={MemGPT: Towards LLMs as Operating Systems}, 
      author={Charles Packer and Sarah Wooders and Kevin Lin and Vivian Fang and Shishir G. Patil and Ion Stoica and Joseph E. Gonzalez},
      year={2024},
      eprint={2310.08560},
      archivePrefix={arXiv},
      primaryClass={cs.AI},
      url={https://arxiv.org/abs/2310.08560}, 
}

@article{chhikara2025mem0,
    author = {Chhikara, Prateek and Khant, Dev and Aryan, Saket and Singh, Taranjeet and Yadav, Deshraj},
    journal = {ArXiv preprint},
    title = {Mem0: Building production-ready ai agents with scalable long-term memory},
    url = {https://arxiv.org/abs/2504.19413},
    volume = {abs/2504.19413},
    year = {2025}
}

@inproceedings{zheng2024synapse,
    author = {Longtao Zheng and
Rundong Wang and
Xinrun Wang and
Bo An},
    bibsource = {dblp computer science bibliography, https://dblp.org},
    booktitle = {The Twelfth International Conference on Learning Representations,
{ICLR} 2024, Vienna, Austria, May 7-11, 2024},
    publisher = {OpenReview.net},
    title = {Synapse: Trajectory-as-Exemplar Prompting with Memory for Computer
Control},
    url = {https://openreview.net/forum?id=Pc8AU1aF5e},
    year = {2024}
}

@inproceedings{wang2025agent,
    author = {Zora Zhiruo Wang and Jiayuan Mao and Daniel Fried and Graham Neubig},
    booktitle = {Forty-second International Conference on Machine Learning},
    title = {Agent Workflow Memory},
    url = {https://openreview.net/forum?id=NTAhi2JEEE},
    year = {2025}
}

@inproceedings{zhou2024webarena,
    author = {Shuyan Zhou and
Frank F. Xu and
Hao Zhu and
Xuhui Zhou and
Robert Lo and
Abishek Sridhar and
Xianyi Cheng and
Tianyue Ou and
Yonatan Bisk and
Daniel Fried and
Uri Alon and
Graham Neubig},
    bibsource = {dblp computer science bibliography, https://dblp.org},
    booktitle = {The Twelfth International Conference on Learning Representations,
{ICLR} 2024, Vienna, Austria, May 7-11, 2024},
    publisher = {OpenReview.net},
    title = {WebArena: {A} Realistic Web Environment for Building Autonomous Agents},
    url = {https://openreview.net/forum?id=oKn9c6ytLx},
    year = {2024}
}

@article{wang2024survey,
    author = {Wang, Lei and Ma, Chen and Feng, Xueyang and Zhang, Zeyu and Yang, Hao and Zhang, Jingsen and Chen, Zhiyuan and Tang, Jiakai and Chen, Xu and Lin, Yankai and others},
    journal = {Frontiers of Computer Science},
    number = {6},
    pages = {186345},
    publisher = {Springer},
    title = {A survey on large language model based autonomous agents},
    volume = {18},
    year = {2024}
}

@inproceedings{hu-etal-2025-hiagent,
    address = {Vienna, Austria},
    author = {Hu, Mengkang  and
Chen, Tianxing  and
Chen, Qiguang  and
Mu, Yao  and
Shao, Wenqi  and
Luo, Ping},
    booktitle = {Proceedings of the 63rd Annual Meeting of the Association for Computational Linguistics (Volume 1: Long Papers)},
    doi = {10.18653/v1/2025.acl-long.1575},
    editor = {Che, Wanxiang  and
Nabende, Joyce  and
Shutova, Ekaterina  and
Pilehvar, Mohammad Taher},
    isbn = {979-8-89176-251-0},
    pages = {32779--32798},
    publisher = {Association for Computational Linguistics},
    title = {{H}i{A}gent: Hierarchical Working Memory Management for Solving Long-Horizon Agent Tasks with Large Language Model},
    url = {https://aclanthology.org/2025.acl-long.1575/},
    year = {2025}
}

@inproceedings{yao2022webshop,
 author = {Yao, Shunyu and Chen, Howard and Yang, John and Narasimhan, Karthik},
 booktitle = {NeurIPS},
 title = {{WebShop}: Towards Scalable Real-World Web Interaction with Grounded Language Agents},
 year = {2022}
}

@misc{webagentr1,
      title={WebAgent-R1: Training Web Agents via End-to-End Multi-Turn Reinforcement Learning}, 
      author={Zhepei Wei and Wenlin Yao and Yao Liu and Weizhi Zhang and Qin Lu and Liang Qiu and Changlong Yu and Puyang Xu and Chao Zhang and Bing Yin and Hyokun Yun and Lihong Li},
      year={2025},
      eprint={2505.16421},
      archivePrefix={arXiv},
      primaryClass={cs.CL},
      url={https://arxiv.org/abs/2505.16421}, 
}

@misc{searchr1,
      title={Search-R1: Training LLMs to Reason and Leverage Search Engines with Reinforcement Learning}, 
      author={Bowen Jin and Hansi Zeng and Zhenrui Yue and Jinsung Yoon and Sercan Arik and Dong Wang and Hamed Zamani and Jiawei Han},
      year={2025},
      eprint={2503.09516},
      archivePrefix={arXiv},
      primaryClass={cs.CL},
      url={https://arxiv.org/abs/2503.09516}, 
}

@misc{reasoningbank,
      title={ReasoningBank: Scaling Agent Self-Evolving with Reasoning Memory}, 
      author={Siru Ouyang and Jun Yan and I-Hung Hsu and Yanfei Chen and Ke Jiang and Zifeng Wang and Rujun Han and Long T. Le and Samira Daruki and Xiangru Tang and Vishy Tirumalashetty and George Lee and Mahsan Rofouei and Hangfei Lin and Jiawei Han and Chen-Yu Lee and Tomas Pfister},
      year={2025},
      eprint={2509.25140},
      archivePrefix={arXiv},
      primaryClass={cs.AI},
      url={https://arxiv.org/abs/2509.25140}, 
}

@inproceedings{TriviaQA,
  author       = {Mandar Joshi and
                  Eunsol Choi and
                  Daniel S. Weld and
                  Luke Zettlemoyer},
  editor       = {Regina Barzilay and
                  Min{-}Yen Kan},
  title        = {TriviaQA: {A} Large Scale Distantly Supervised Challenge Dataset for
                  Reading Comprehension},
  booktitle    = {Proceedings of the 55th Annual Meeting of the Association for Computational
                  Linguistics, {ACL} 2017, Vancouver, Canada, July 30 - August 4, Volume
                  1: Long Papers},
  pages        = {1601--1611},
  publisher    = {Association for Computational Linguistics},
  year         = {2017},
  url          = {https://doi.org/10.18653/v1/P17-1147},
  doi          = {10.18653/V1/P17-1147},
  bibsource    = {dblp computer science bibliography, https://dblp.org}
}

@article{NQ,
  author       = {Tom Kwiatkowski and
                  Jennimaria Palomaki and
                  Olivia Redfield and
                  Michael Collins and
                  Ankur P. Parikh and
                  Chris Alberti and
                  Danielle Epstein and
                  Illia Polosukhin and
                  Jacob Devlin and
                  Kenton Lee and
                  Kristina Toutanova and
                  Llion Jones and
                  Matthew Kelcey and
                  Ming{-}Wei Chang and
                  Andrew M. Dai and
                  Jakob Uszkoreit and
                  Quoc Le and
                  Slav Petrov},
  title        = {Natural Questions: a Benchmark for Question Answering Research},
  journal      = {Trans. Assoc. Comput. Linguistics},
  volume       = {7},
  pages        = {452--466},
  year         = {2019},
  url          = {https://doi.org/10.1162/tacl\_a\_00276},
  doi          = {10.1162/TACL\_A\_00276},
  bibsource    = {dblp computer science bibliography, https://dblp.org}
}

@article{PopQA,
  author       = {Alex Mallen and
                  Akari Asai and
                  Victor Zhong and
                  Rajarshi Das and
                  Hannaneh Hajishirzi and
                  Daniel Khashabi},
  title        = {When Not to Trust Language Models: Investigating Effectiveness and
                  Limitations of Parametric and Non-Parametric Memories},
  journal      = {CoRR},
  volume       = {abs/2212.10511},
  year         = {2022},
  url          = {https://doi.org/10.48550/arXiv.2212.10511},
  doi          = {10.48550/ARXIV.2212.10511},
  eprinttype    = {arXiv},
  eprint       = {2212.10511},
  bibsource    = {dblp computer science bibliography, https://dblp.org}
}

@inproceedings{HotpotQA,
  author       = {Zhilin Yang and
                  Peng Qi and
                  Saizheng Zhang and
                  Yoshua Bengio and
                  William W. Cohen and
                  Ruslan Salakhutdinov and
                  Christopher D. Manning},
  editor       = {Ellen Riloff and
                  David Chiang and
                  Julia Hockenmaier and
                  Jun'ichi Tsujii},
  title        = {HotpotQA: {A} Dataset for Diverse, Explainable Multi-hop Question
                  Answering},
  booktitle    = {Proceedings of the 2018 Conference on Empirical Methods in Natural
                  Language Processing, Brussels, Belgium, October 31 - November 4, 2018},
  pages        = {2369--2380},
  publisher    = {Association for Computational Linguistics},
  year         = {2018},
  url          = {https://doi.org/10.18653/v1/d18-1259},
  doi          = {10.18653/V1/D18-1259},
  bibsource    = {dblp computer science bibliography, https://dblp.org}
}

@article{Musique,
  author       = {Harsh Trivedi and
                  Niranjan Balasubramanian and
                  Tushar Khot and
                  Ashish Sabharwal},
  title        = {MuSiQue: Multihop Questions via Single-hop Question
                  Composition},
  journal      = {Trans. Assoc. Comput. Linguistics},
  volume       = {10},
  pages        = {539--554},
  year         = {2022},
  url          = {https://doi.org/10.1162/tacl\_a\_00475},
  doi          = {10.1162/TACL\_A\_00475},
  bibsource    = {dblp computer science bibliography, https://dblp.org}
}

@inproceedings{Bamboogle,
  author       = {Ofir Press and
                  Muru Zhang and
                  Sewon Min and
                  Ludwig Schmidt and
                  Noah A. Smith and
                  Mike Lewis},
  editor       = {Houda Bouamor and
                  Juan Pino and
                  Kalika Bali},
  title        = {Measuring and Narrowing the Compositionality Gap in Language Models},
  booktitle    = {Findings of the Association for Computational Linguistics: {EMNLP}
                  2023, Singapore, December 6-10, 2023},
  pages        = {5687--5711},
  publisher    = {Association for Computational Linguistics},
  year         = {2023},
  url          = {https://doi.org/10.18653/v1/2023.findings-emnlp.378},
  doi          = {10.18653/V1/2023.FINDINGS-EMNLP.378},
  bibsource    = {dblp computer science bibliography, https://dblp.org}
}

@article{Search-R1,
  author       = {Bowen Jin and
                  Hansi Zeng and
                  Zhenrui Yue and
                  Dong Wang and
                  Hamed Zamani and
                  Jiawei Han},
  title        = {Search-R1: Training LLMs to Reason and Leverage Search Engines with
                  Reinforcement Learning},
  journal      = {CoRR},
  volume       = {abs/2503.09516},
  year         = {2025},
  url          = {https://doi.org/10.48550/arXiv.2503.09516},
  doi          = {10.48550/ARXIV.2503.09516},
  eprinttype    = {arXiv},
  eprint       = {2503.09516},
  bibsource    = {dblp computer science bibliography, https://dblp.org}
}

@article{gpt4o,
  title={Gpt-4o system card},
  author={Hurst, Aaron and Lerer, Adam and Goucher, Adam P and Perelman, Adam and Ramesh, Aditya and Clark, Aidan and Ostrow, AJ and Welihinda, Akila and Hayes, Alan and Radford, Alec and others},
  journal={arXiv preprint arXiv:2410.21276},
  year={2024}
}

@article{qwen_2_5,
  author       = {An Yang and
                  Baosong Yang and
                  Beichen Zhang and
                  Binyuan Hui and
                  Bo Zheng and
                  Bowen Yu and
                  Chengyuan Li and
                  Dayiheng Liu and
                  Fei Huang and
                  Haoran Wei and
                  Huan Lin and
                  Jian Yang and
                  Jianhong Tu and
                  Jianwei Zhang and
                  Jianxin Yang and
                  Jiaxi Yang and
                  Jingren Zhou and
                  Junyang Lin and
                  Kai Dang and
                  Keming Lu and
                  Keqin Bao and
                  Kexin Yang and
                  Le Yu and
                  Mei Li and
                  Mingfeng Xue and
                  Pei Zhang and
                  Qin Zhu and
                  Rui Men and
                  Runji Lin and
                  Tianhao Li and
                  Tingyu Xia and
                  Xingzhang Ren and
                  Xuancheng Ren and
                  Yang Fan and
                  Yang Su and
                  Yichang Zhang and
                  Yu Wan and
                  Yuqiong Liu and
                  Zeyu Cui and
                  Zhenru Zhang and
                  Zihan Qiu},
  title        = {Qwen2.5 Technical Report},
  journal      = {CoRR},
  volume       = {abs/2412.15115},
  year         = {2024},
  url          = {https://doi.org/10.48550/arXiv.2412.15115},
  doi          = {10.48550/ARXIV.2412.15115},
  eprinttype    = {arXiv},
  eprint       = {2412.15115},
  bibsource    = {dblp computer science bibliography, https://dblp.org}
}

@article{huang2025language,
  title={Language Models Do Not Have Human-Like Working Memory},
  author={Huang, Jen-tse and Sun, Kaiser and Wang, Wenxuan and Dredze, Mark},
  journal={arXiv preprint arXiv:2505.10571},
  year={2025}
}

@article{du2025rethinking,
  title={Rethinking memory in ai: Taxonomy, operations, topics, and future directions},
  author={Du, Yiming and Huang, Wenyu and Zheng, Danna and Wang, Zhaowei and Montella, Sebastien and Lapata, Mirella and Wong, Kam-Fai and Pan, Jeff Z},
  journal={arXiv e-prints},
  pages={arXiv--2505},
  year={2025}
}

@misc{langgraph,
  author = {LangChain},
  title = {LangGraph: Building Stateful, Multi-Actor Applications with LLMs},
  year = {2024},
  publisher = {GitHub},
  journal = {GitHub repository},
  howpublished = {\url{https://github.com/langchain-ai/langgraph}}
}

@article{jiang2026magma,
  title={MAGMA: A Multi-Graph based Agentic Memory Architecture for AI Agents},
  author={Jiang, Dongming and Li, Yi and Li, Guanpeng and Li, Bingzhe},
  journal={arXiv preprint arXiv:2601.03236},
  year={2026}
}

@inproceedings{ming2025anchorcot,
  title={AnchorCoT: Anchors Pave the Way for Multi-hop Reasoning},
  author={Ming, Tianshi and Wu, Xian and Zhang, Yingying and Fu, Zichuan and Cheng, Dawei},
  booktitle={Findings of the Association for Computational Linguistics: ACL 2025},
  pages={15522--15536},
  year={2025}
}

@article{yan2025memory,
  title={Memory-r1: Enhancing large language model agents to manage and utilize memories via reinforcement learning},
  author={Yan, Sikuan and Yang, Xiufeng and Huang, Zuchao and Nie, Ercong and Ding, Zifeng and Li, Zonggen and Ma, Xiaowen and Bi, Jinhe and Kersting, Kristian and Pan, Jeff Z and others},
  journal={arXiv preprint arXiv:2508.19828},
  year={2025}
}

@article{zhang2025unlocking,
  title={Unlocking the Power of Multi-Agent LLM for Reasoning: From Lazy Agents to Deliberation},
  author={Zhang, Zhiwei and Li, Xiaomin and Lin, Yudi and Liu, Hui and Chandradevan, Ramraj and Wu, Linlin and Lin, Minhua and Wang, Fali and Tang, Xianfeng and He, Qi and others},
  journal={arXiv preprint arXiv:2511.02303},
  year={2025}
}

@misc{25tfgrpo,
      title={Training-Free Group Relative Policy Optimization}, 
      author={Yuzheng Cai and Siqi Cai and Yuchen Shi and Zihan Xu and Lichao Chen and Yulei Qin and Xiaoyu Tan and Gang Li and Zongyi Li and Haojia Lin and Yong Mao and Ke Li and Xing Sun},
      year={2025},
      eprint={2510.08191},
      archivePrefix={arXiv},
      primaryClass={cs.CL},
      url={https://arxiv.org/abs/2510.08191}, 
}
\bibliographystyle{colm2026_conference}

\newpage
\appendix
\onecolumn

\section{Context Calculation}\label{sec:context_calculation}

Effective context management is paramount for long-horizon tasks in web and search environments, where agents interact through multiple turns and observations can be highly dimensional. This section details how the context length is calculated in both the WebArena and DeepSearch tasks under three distinct memory configurations: \textit{No Memory}, \textit{Full Context}, and our proposed \textit{ActiveContext}.

We define the context length for a single turn $t$ as $C_t$. The total context length for a task spanning $L$ turns is then $\sum_{t=1}^{L} C_t$.

\subsection{No Memory}
The \textit{No Memory} configuration streamlines the agent's input at each turn by restricting the conversational context to only immediate necessities, completely discarding prior interaction history.

For web exploration tasks, the input at turn $t$ is limited to the system prompt ($S$), the current web XTREE observation ($O_t^{\text{XTREE}}$), and the task objective ($U$). The context length for turn $t$ under the \textit{No Memory} configuration, $C_t^{\text{NoMem}}$, is formally expressed as:
\begin{equation}
    C_t^{\text{NoMem}} = \text{len}(S) + \text{len}(O_t^{\text{XTREE}}) + \text{len}(U)
\end{equation}

While this approach significantly reduces the per-turn context length compared to \textit{Full Context}, it often leads to a loss of crucial long-term dependencies and sequential understanding necessary for complex, multi-step tasks. 

Notably, for Search tasks, a \textit{No Memory} configuration was not evaluated due to the prevalence of multi-hop search requirements, which are inherently impossible to solve without historical context. In contrast, Web tasks can still proceed under a \textit{No Memory} setup, as the agent can often infer subsequent operations based purely on the current visual state of the web observation and the global objective.

\subsection{Full Context}
In the \textit{Full Context} configuration, the agent's input at each turn $t$ comprises the cumulative historical conversation. This includes the initial system prompt, user prompts (observations and objectives), and assistant prompts (reasoning and actions). 

\begin{itemize}[leftmargin=*]
    \item \textbf{Web Tasks:} To prevent catastrophic context bloat from massive DOM trees, historical observations are replaced with a concise placeholder ($P$). The input at turn $t$ includes the system prompt ($S$), the \textit{current} observation ($O_t^{\text{AAT}}$), placeholders for all past observations ($\sum_{k=0}^{t-1} P$), the task objective ($U$), and all prior assistant prompts detailing reasoning ($Re_k$) and actions ($A_k$).
    \item \textbf{Search Tasks:} Because text retrieval is less structurally verbose than DOM trees, placeholders are not used. The input includes the system prompt ($S$), the entire history of retrieved information ($\sum_{k=0}^{t} R_k$), the final objective ($U$), and all previous assistant prompts ($\sum_{k=0}^{t-1} (Re_k, A_k)$).
\end{itemize}

The context length for turn $t$ under the \textit{Full Context} configuration, $C_t^{\text{FullContext}}$, is formally expressed as:
\begin{equation}
    C_t^{\text{FullContext}} = \text{len}(S) + \text{len}(O_t^{\text{AAT}}) + \sum_{k=0}^{t-1} \text{len}(P) + \text{len}(U) + \sum_{k=0}^{t-1} \text{len}(Re_k, A_k) \quad \text{(Web)}
\end{equation}
\begin{equation}
    C_t^{\text{FullContext}} = \text{len}(S) + \sum_{k=0}^{t} \text{len}(R_k) + \text{len}(U) + \sum_{k=0}^{t-1} \text{len}(Re_k, A_k) \quad \text{(Search)}
\end{equation}

A significant challenge with \textit{Full Context} is that the input context length grows linearly with the number of interaction turns $t$. This cumulative growth inevitably triggers the "Context Bottleneck" described previously, significantly increasing computational costs and exacerbating the "Lost-in-the-Middle" phenomenon, thereby severely impeding reasoning capabilities.

\subsection{ActiveContext}
Under the \textit{ActiveContext} framework, the TaskExecutor's input at each turn $t$ is augmented with a refined working memory $\mathcal{M}_t$ dynamically generated by the ContextCurator. This module is specifically designed to condense salient information from the interaction history into a highly compact and relevant representation.

\begin{itemize}[leftmargin=*]
    \item \textbf{Web Tasks:} The TaskExecutor's input includes the system prompt ($S$), the current web XTREE observation ($O_t^{\text{XTREE}}$), the objective ($U$), and crucially, the refined memory ($\mathcal{M}_t^{\text{Web}}$). This memory represents a distilled summary of past interactions and reasoning anchors, rather than a raw, concatenated history.
    \item \textbf{Search Tasks:} The TaskExecutor's input includes the system prompt ($S$), the just-retrieved search results ($R_t$), the objective ($U$), and the refined memory ($\mathcal{M}_t^{\text{Search}}$) containing historical search content, key retrieved evidence, and past erroneous deductions.
\end{itemize}

The context length for turn $t$ under the \textit{ActiveContext} configuration, $C_t^{\text{Active}}$, is formally expressed as:
\begin{equation}
    C_t^{\text{Active}} = \text{len}(S) + \text{len}(O_t^{\text{XTREE}}) + \text{len}(U) + \text{len}(\mathcal{M}_t^{\text{Web}}) \quad \text{(Web)}
\end{equation}
\begin{equation}
    C_t^{\text{Active}} = \text{len}(S) + \text{len}(R_t) + \text{len}(U) + \text{len}(\mathcal{M}_t^{\text{Search}}) \quad \text{(Search)}
\end{equation}

This symbiotic approach strikes an optimal balance: it halts the linear growth of context length observed in the \textit{Full Context} baseline while actively retaining essential historical anchors. By aggressively filtering noise, it mitigates the context bottleneck and maximizes the signal-to-noise ratio in the TaskExecutor's decision-making process. Although the per-turn input might occasionally be slightly longer than in the \textit{No Memory} setup, \textit{ActiveContext} achieves a substantially higher task success rate and requires fewer exploratory interaction turns per task. This ultimately results in a significantly shorter total context sequence required for successful task completion, demonstrating superior computational efficiency and reasoning effectiveness.

%%%%%%%%%%%%%%%%%%%%%%%%%%%%%%%%%%%%%%%%%%%%%%%%%%%%%%%%%%%%%%%%%%%%%%%%%%%%%%%
%%%%%%%%%%%%%%%%%%%%%%%%%%%%%%%%%%%%%%%%%%%%%%%%%%%%%%%%%%%%%%%%%%%%%%%%%%%%%%%

\section{Case Study: Multi-turn Interaction with ActiveContext}\label{sec:case_study}

To illustrate the effectiveness of our \textit{ActiveContext} framework, we present a detailed case study of a multi-turn interaction from the WebArena benchmark. This example demonstrates how the \texttt{gpt-4o-mini} model, acting as the \textit{TaskExecutor} and equipped with a RL-trained \textit{ContextCurator}, navigates a complex web environment to achieve a specific objective, effectively managing context and refining its working memory.

\subsection{Task Objective}
The agent's objective for this task is: \textit{``Tell me the count of comments that have received more downvotes than upvotes for the user who made the latest post on the Showerthoughts forum."}

\subsection{Interaction Log Analysis}
The following sequence details the agent's steps, its reasoning, and the dynamically refined memory provided by the \textit{ContextCurator}. We highlight two critical, consecutive turns (Turn 7 and Turn 8) in the interaction to demonstrate the information entropy reduction.

\begin{tcolorbox}[breakable,colback=gray!5!white,colframe=gray!50!black,title=\textbf{Turn 7: Navigating to the Comments Section}]
\textbf{1. ContextCurator (Memory Update):} \\
The WebAgent identified `MarvelsGrantMan136' as the user who made the latest post under the `Postmill' tab. This post had no comments. The agent then navigated to `MarvelsGrantMan136's profile.

Initial traversal of the user's profile and previous posts in the `Postmill' tab did not yield relevant results regarding comments or vote counts, except for one old comment with a single downvote. To view all comments, the WebAgent clicked on the `Comments' link on the user's profile page.

The next page loaded, displaying a list of comments made by `MarvelsGrantMan136'. The current page structure, as observed:

- Main content area for `MarvelsGrantMan136'

- Navigation links: `Overview', `Submissions', `Comments' (link ID: [3018])

- An article section, potentially containing comment details or links to posts that have comments.

\vspace{2mm}\noindent\tikz\draw[dashed, gray!50!black] (0,0) -- (\linewidth,0);\vspace{2mm}

\textbf{2. Latest Web Observation:} \\
{\scriptsize \ttfamily
Tab 0 (current): MarvelsGrantMan136

\quad[2008] RootWebArea `MarvelsGrantMan136' focused: True

\quad\quad[2287] HeaderAsNonLandmark 

\quad\quad\quad[2288] link `Home'

\quad\quad\quad\quad[2334] StaticText `Postmill'

\quad\quad[2335] link `Forums'

\quad\quad[2336] link `Wiki'

\quad\quad[2344] searchbox `Search query'

\quad\quad[2345] link `Notifications (0)'

\quad\quad[2346] link `Submit'

\quad\quad[2019] button `MarvelsGrantMan136' hasPopup: menu expanded: False

\quad\quad[2021] main

\quad\quad\quad[3016] link `Overview'

\quad\quad\quad[3017] link `Submissions'

\quad\quad\quad[3018] link `Comments'

\quad\quad\quad[2022] article 

\quad\quad\quad\quad[2025] HeaderAsNonLandmark

\quad\quad\quad\quad\quad[2354] heading `How Machine Learning Can Revolutionize the Book Industry'

\quad\quad\quad\quad\quad\quad[3019] link `How Machine Learning Can Revolutionize the Book Industry'

\quad\quad\quad\quad\quad[3021] StaticText `Submitted by '

\quad\quad\quad\quad\quad[2028] link `MarvelsGrantMan136' expanded: False

\quad\quad\quad\quad\quad[3023] time `January 13, 2026 at 12:55:18 PM EST'

\quad\quad\quad\quad\quad\quad[3609] StaticText `16 hours ago'

\quad\quad\quad\quad\quad[3024] StaticText ` in '

\quad\quad\quad\quad\quad[3025] link `books'

\quad\quad\quad\quad[3612] link `No comments'

\quad\quad\quad\quad[3614] link `Edit'

\quad\quad\quad\quad[4077] button `Delete'

\quad\quad\quad\quad[2358] button `Retract upvote'

\quad\quad\quad\quad[2361] button `Downvote'

\quad\quad\quad[2029] article 

\quad\quad\quad\quad[2303] checkbox `Hide \textminus' checked: true

\quad\quad\quad\quad[2032] HeaderAsNonLandmark

\quad\quad\quad\quad\quad[2033] heading `MarvelsGrantMan136 wrote 16 hours ago'

\quad\quad\quad\quad\quad\quad[2035] link `MarvelsGrantMan136' expanded: False

\quad\quad\quad\quad\quad\quad[3039] time `January 13, 2026 at 12:51:42 PM EST'

\quad\quad\quad\quad[2376] StaticText `Reply to '

\quad\quad\quad\quad[2377] link `AI experts disown Musk-backed campaign citing their research'

\quad\quad\quad\quad[2378] StaticText ` by '

\quad\quad\quad\quad[2039] link `Don\_Gato1' expanded: False

\quad\quad\quad\quad[2380] StaticText 

\quad\quad\quad\quad[3628] link `Reply'

\quad\quad\quad\quad[3629] link `Permalink'

\quad\quad\quad\quad[4082] button `Delete'

\quad\quad\quad\quad[3631] link `Edit'

\quad\quad\quad\quad[2384] button `Retract upvote'

\quad\quad\quad\quad[2387] button `Downvote'

\quad\quad\quad[2040] article 

\quad\quad\quad\quad[2043] HeaderAsNonLandmark 

\quad\quad\quad\quad\quad[2391] heading `How Machine Learning Can Transform the Corresponding Field How Machine Learning Can Transform the Corresponding Field'

\quad\quad\quad\quad\quad\quad[3060] link `How Machine Learning Can Transform the Corresponding Field How Machine Learning Can Transform the Corresponding Field'

\quad\quad\quad\quad\quad[3062] StaticText `Submitted by '

\quad\quad\quad\quad\quad[2046] link `MarvelsGrantMan136' expanded: False

\quad\quad\quad\quad\quad[3064] time `January 13, 2026 at 9:09:04 AM EST'

\quad\quad\quad\quad\quad\quad[3646] StaticText `20 hours ago'

\quad\quad\quad\quad\quad[3065] StaticText ` in '

\quad\quad\quad\quad\quad[3066] link `books'

\quad\quad\quad\quad[3649] link `No comments'

\quad\quad\quad\quad[3651] link `Edit'

\quad\quad\quad\quad[4088] button `Delete'

\quad\quad\quad\quad[2395] button `Upvote'

\quad\quad\quad\quad[2398] button `Downvote'

\quad\quad\quad[2047] article 

\quad\quad\quad\quad[2304] checkbox `Hide \textminus' checked: true

\quad\quad\quad\quad[2050] HeaderAsNonLandmark 

\quad\quad\quad\quad\quad[2051] heading `MarvelsGrantMan136 wrote 20 hours ago'

\quad\quad\quad\quad\quad\quad[2053] link `MarvelsGrantMan136' expanded: False

\quad\quad\quad\quad\quad\quad[3080] time `January 13, 2026 at 9:05:14 AM EST'

\quad\quad\quad\quad[2413] StaticText `Reply to '

\quad\quad\quad\quad[2414] link `AI experts disown Musk-backed campaign citing their research'

\quad\quad\quad\quad[2415] StaticText ` by '

\quad\quad\quad\quad[2057] link `Don\_Gato1' expanded: False

\quad\quad\quad\quad[2417] StaticText

\quad\quad\quad\quad[3665] link `Reply'

\quad\quad\quad\quad[3666] link `Permalink'

\quad\quad\quad\quad[4093] button `Delete'

\quad\quad\quad\quad[3668] link `Edit'

\quad\quad\quad\quad[2421] button `Retract upvote'

\quad\quad\quad\quad[2424] button `Downvote'

\quad\quad\quad[2058] article

\quad\quad\quad\quad[2305] checkbox `Hide \textminus' checked: true

\quad\quad\quad\quad[2061] HeaderAsNonLandmark 

\quad\quad\quad\quad\quad[2062] heading `MarvelsGrantMan136 wrote 23 hours ago'

\quad\quad\quad\quad\quad\quad[2064] link `MarvelsGrantMan136' expanded: False

\quad\quad\quad\quad\quad\quad[3102] time `January 13, 2026 at 5:59:50 AM EST'

\quad\quad\quad\quad[2439] StaticText `Reply to '

\quad\quad\quad\quad[2440] link `Driving in Pittsburgh summed up by one traffic sign.'

\quad\quad\quad\quad[2068] link `Dare\_County' expanded: False

\quad\quad\quad\quad[2443] StaticText `Yeah, pittsburgh traffic, you know...Yeah, pittsburgh traffic, you know...'

\quad\quad\quad\quad[3681] link `Reply'

\quad\quad\quad\quad[3682] link `Permalink'

\quad\quad\quad\quad[4099] button `Delete'

\quad\quad\quad\quad[3684] link `Edit'

\quad\quad\quad\quad[2447] button `Retract upvote'

\quad\quad\quad\quad[3115] StaticText `1'

\quad\quad\quad\quad[2450] button `Downvote'

\quad\quad[2277] complementary 

\quad\quad\quad[2322] HeaderAsNonLandmark 

\quad\quad\quad\quad[2902] heading `MarvelsGrantMan136'

\quad\quad\quad\quad\quad[3574] link `MarvelsGrantMan136'

\quad\quad\quad\quad[3577] StaticText `Registered '

\quad\quad\quad\quad[3578] time `June 24, 2020 at 11:02:10 AM EDT'

\quad\quad\quad\quad\quad[4061] StaticText `6 years ago'

\quad\quad\quad\quad[3581] StaticText `Not whitelisted'

\quad\quad\quad[2905] StaticText `t2\_5adwlxvnAwesome Prompt ArtistFreelance Web DeveloperAwesome Prompt ArtistAwesome Prompt ArtistAwesome Prompt ArtistFreelance Web DeveloperI am a robotI am a robotPro Python Developer withFreelance Web Developer Freelance Web Developer Freelance Web Developer 20 years of Experience'

\quad\quad\quad[2906] link `Edit biography'

\quad\quad\quad[2325] heading `Toolbox'

\quad\quad\quad[3586] link `Hidden forums'

\quad\quad\quad[3587] link `Trash'

\quad\quad\quad[2327] heading `MarvelsGrantMan136 is a moderator on:'

\quad\quad\quad[3588] link `PlantsForCatParents'

\quad\quad\quad[3589] link `cmu\_lti'

\quad\quad\quad[3590] link `KaraokeKaraoke'

\quad\quad\quad[3591] link `Karaoke'

\quad\quad\quad[3592] link `sci\_fi'

\quad\quad\quad[3593] link `cmu\_lticmu\_lti'

URL: \url{http://3.151.53.169:9999/user/MarvelsGrantMan136}
}

\vspace{2mm}\noindent\tikz\draw[dashed, gray!50!black] (0,0) -- (\linewidth,0);\vspace{2mm}

\textbf{3. TaskExecutor's Reasoning and Action (\texttt{gpt-4o-mini}):} \\
Let's think step-by-step. According to the memory management agent, I have navigated to the comments section for the user `MarvelsGrantMan136'. The previous action was to click on the `Comments' link, which should have led to a list of comments made by this user. However, the observation does not provide specific details about the comments or their vote counts.
    
Since I need to find out how many comments have received more downvotes than upvotes, I should look for the relevant information regarding the comments made by `MarvelsGrantMan136'. The next logical step is to ensure that I am viewing the comments section properly and to check if there are any comments listed that indicate their vote counts.

In summary, the next action I will perform is ```click [3017]''' to access the comments made by `MarvelsGrantMan136'.
\end{tcolorbox}

\begin{tcolorbox}[breakable,colback=gray!5!white,colframe=gray!50!black,title=\textbf{Turn 8: Task Completion}]
\textbf{1. ContextCurator (Memory Update):} \\
The WebAgent initially found that 'MarvelsGrantMan136' made the latest post under the `Postmill' tab, which had no comments. The agent then traversed the user's profile to view posts and comments. After viewing the profile, the `Comments' link was clicked to see all comments made by `MarvelsGrantMan136'.

The page then loaded a list of comments from `MarvelsGrantMan136'. Subsequently, the WebAgent clicked on the `Comments' link (ID [3017]) to ensure proper navigation within the comments section.

\vspace{2mm}\noindent\tikz\draw[dashed, gray!50!black] (0,0) -- (\linewidth,0);\vspace{2mm}

\textbf{2. Latest Web Observation:} \\
{\scriptsize \ttfamily
    Tab 0 (current): Submissions

    \quad[4254] RootWebArea `Submissions' focused: True
    
    \quad\quad[4458] HeaderAsNonLandmark 
    
    \quad\quad\quad[4459] link `Home'
    
    \quad\quad\quad\quad[4487] StaticText `Postmill'
    
    \quad\quad[4488] link `Forums'
    
    \quad\quad[4489] link `Wiki'
    
    \quad\quad[4497] searchbox `Search query'
    
    \quad\quad[4498] link `Notifications (0)'
    
    \quad\quad[4499] link `Submit'
    
    \quad\quad[4265] button `MarvelsGrantMan136' hasPopup: menu expanded: False
    
    \quad\quad[4267] main 
    
    \quad\quad\quad[4946] link `Overview'
    
    \quad\quad\quad[4947] link `Submissions'
    
    \quad\quad\quad[4948] link `Comments'
    
    \quad\quad\quad[4268] article 
    
    \quad\quad\quad\quad[4271] HeaderAsNonLandmark 
    
    \quad\quad\quad\quad\quad[4507] heading `The Last Of Us’ Moves Production to Vancouver For Season 2
    
    \quad\quad\quad\quad\quad\quad[4949] link `The Last Of Us’ Moves Production to Vancouver For Season 2
    
    \quad\quad\quad\quad\quad[4509] link `deadline.com'
    
    \quad\quad\quad\quad\quad[4952] StaticText `Submitted by '
    
    \quad\quad\quad\quad\quad[4274] link `MarvelsGrantMan136' expanded: False
    
    \quad\quad\quad\quad\quad[5536] StaticText `t3\_127rvdp'
    
    \quad\quad\quad\quad\quad[4956] time `March 31, 2023 at 1:02:17 PM EDT'
    
    \quad\quad\quad\quad\quad\quad[5540] StaticText `3 years ago'
    
    \quad\quad\quad\quad\quad[4957] StaticText ` in '
    
    \quad\quad\quad\quad\quad[4958] link `television'
    
    \quad\quad\quad\quad[5543] link `No comments'
    
    \quad\quad\quad\quad[5545] link `Edit'
    
    \quad\quad\quad\quad[6200] button `Delete'
    
    \quad\quad\quad\quad[4513] button `Upvote'
    
    \quad\quad\quad\quad[4516] button `Downvote'
    
    \quad\quad\quad[4275] article 
    
    \quad\quad\quad\quad[4278] HeaderAsNonLandmark 
    
    \quad\quad\quad\quad\quad[4520] heading ```Secret Invasion' Revealed: Inside Samuel L. Jackson’s Eye-Opening New Marvel Series"
    
    \quad\quad\quad\quad\quad\quad[4971] link ```Secret Invasion' Revealed: Inside Samuel L. Jackson’s Eye-Opening New Marvel Series"
    
    \quad\quad\quad\quad\quad[4522] link `vanityfair.com'
    
    \quad\quad\quad\quad\quad[4974] StaticText `Submitted by '
    
    \quad\quad\quad\quad\quad[4281] link `MarvelsGrantMan136' expanded: False
    
    \quad\quad\quad\quad\quad[5562] StaticText `t3\_127jaxx'
    
    \quad\quad\quad\quad\quad[4978] time `March 31, 2023 at 8:12:27 AM EDT'
    
    \quad\quad\quad\quad\quad\quad[5566] StaticText `3 years ago'
    
    \quad\quad\quad\quad\quad[4979] StaticText ` in '
    
    \quad\quad\quad\quad\quad[4980] link `television'
    
    \quad\quad\quad\quad[5569] link `No comments'
    
    \quad\quad\quad\quad[5571] link `Edit'
    
    \quad\quad\quad\quad[6205] button `Delete'
    
    \quad\quad\quad\quad[4526] button `Upvote'
    
    \quad\quad\quad\quad[4529] button `Downvote'
    
    \quad\quad\quad[4282] article 
    
    \quad\quad\quad\quad[4285] HeaderAsNonLandmark 
    
    \quad\quad\quad\quad\quad[4533] heading `Star Trek: Starfleet Academy Series, From Alex Kurtzman and Nancy Drew Creator, Ordered at Paramount+'
    
    \quad\quad\quad\quad\quad\quad[4993] link `Star Trek: Starfleet Academy Series, From Alex Kurtzman and Nancy Drew Creator, Ordered at Paramount+'
    
    \quad\quad\quad\quad\quad[4535] link `tvline.com'
    
    \quad\quad\quad\quad\quad[4996] StaticText `Submitted by '
    
    \quad\quad\quad\quad\quad[4288] link `MarvelsGrantMan136' expanded: False
    
    \quad\quad\quad\quad\quad[5588] StaticText `t3\_126sut3'
    
    \quad\quad\quad\quad\quad[5000] time `March 30, 2023 at 1:00:27 PM EDT'
    
    \quad\quad\quad\quad\quad\quad[5592] StaticText `3 years ago'
    
    \quad\quad\quad\quad\quad[5001] StaticText ` in '
    
    \quad\quad\quad\quad\quad[5002] link `television'
    
    \quad\quad\quad\quad[5595] link `No comments'
    
    \quad\quad\quad\quad[5597] link `Edit'
    
    \quad\quad\quad\quad[6210] button `Delete'
    
    \quad\quad\quad\quad[4539] button `Upvote'
    
    \quad\quad\quad\quad[4542] button `Downvote'
    
    \quad\quad\quad[4289] article 
    
    \quad\quad\quad\quad[4292] HeaderAsNonLandmark 
    
    \quad\quad\quad\quad\quad[4546] heading ```Fargo' Season 5: FX Unveils First Look at Jon Hamm, Premieres in Late 2023"
    
    \quad\quad\quad\quad\quad\quad[5015] link ```Fargo' Season 5: FX Unveils First Look at Jon Hamm, Premieres in Late 2023"
    
    \quad\quad\quad\quad\quad[4548] link `tvinsider.com'
    
    \quad\quad\quad\quad\quad[5018] StaticText `Submitted by '
    
    \quad\quad\quad\quad\quad[4295] link `MarvelsGrantMan136' expanded: False
    
    \quad\quad\quad\quad\quad[5614] StaticText `t3\_12686kh'
    
    \quad\quad\quad\quad\quad[5022] time `March 29, 2023 at 10:05:47 PM EDT'
    
    \quad\quad\quad\quad\quad\quad[5618] StaticText `3 years ago'
    
    \quad\quad\quad\quad\quad[5023] StaticText ` in '
    
    \quad\quad\quad\quad\quad[5024] link `television'
    
    \quad\quad\quad\quad[5621] link `No comments'
    
    \quad\quad\quad\quad[5623] link `Edit'
    
    \quad\quad\quad\quad[6215] button `Delete'
    
    \quad\quad\quad\quad[4552] button `Upvote'
    
    \quad\quad\quad\quad[5029] StaticText `0'
    
    \quad\quad\quad\quad[4555] button `Downvote'
    
    \quad\quad\quad[4296] article 
    
    \quad\quad\quad\quad[4299] HeaderAsNonLandmark 
    
    \quad\quad\quad\quad\quad[4559] heading `Disney Lays Off Ike Perlmutter, Chairman of Marvel Entertainment'
    
    \quad\quad\quad\quad\quad\quad[5037] link `Disney Lays Off Ike Perlmutter, Chairman of Marvel Entertainment'
    
    \quad\quad\quad\quad\quad[4561] link `nytimes.com'
    
    \quad\quad\quad\quad\quad[5040] StaticText `Submitted by '
    
    \quad\quad\quad\quad\quad[4302] link `MarvelsGrantMan136' expanded: False
    
    \quad\quad\quad\quad\quad[5640] StaticText `t3\_125s0rj'
    
    \quad\quad\quad\quad\quad[5044] time `March 29, 2023 at 11:49:11 AM EDT'
    
    \quad\quad\quad\quad\quad\quad[5644] StaticText `3 years ago'
    
    \quad\quad\quad\quad\quad[5045] StaticText ` in '
    
    \quad\quad\quad\quad\quad[5046] link `television'
    
    \quad\quad\quad\quad[5647] link `No comments'
    
    \quad\quad\quad\quad[5649] link `Edit'
    
    \quad\quad\quad\quad[6220] button `Delete'
    
    \quad\quad\quad\quad[4565] button `Upvote'
    
    \quad\quad\quad\quad[5051] StaticText `3'
    
    \quad\quad\quad\quad[4568] button `Downvote'
    
    \quad\quad\quad[4303] article 
    
    \quad\quad\quad\quad[4306] HeaderAsNonLandmark 
    
    \quad\quad\quad\quad\quad[4572] heading `House Of The Dragon’ To Get Sorter Season 2 (8 Episodes) As HBO Series Eyes Season 3 Greenlight
    
    \quad\quad\quad\quad\quad\quad[5059] link `House Of The Dragon’ To Get Shorter Season 2 (8 Episodes) As HBO Series Eyes Season 3 Greenlight
    
    \quad\quad\quad\quad\quad[4574] link `deadline.com'
    
    \quad\quad\quad\quad\quad[5062] StaticText `Submitted by '
    
    \quad\quad\quad\quad\quad[4309] link `MarvelsGrantMan136' expanded: False
    
    \quad\quad\quad\quad\quad[5666] StaticText `t3\_1257zaj'
    
    \quad\quad\quad\quad\quad[5066] time `March 28, 2023 at 8:42:17 PM EDT'
    
    \quad\quad\quad\quad\quad\quad[5670] StaticText `3 years ago'
    
    \quad\quad\quad\quad\quad[5067] StaticText ` in '
    
    \quad\quad\quad\quad\quad[5068] link `television'
    
    \quad\quad\quad\quad[5673] link `No comments'
    
    \quad\quad\quad\quad[5675] link `Edit'
    
    \quad\quad\quad\quad[6225] button `Delete'
    
    \quad\quad\quad\quad[4578] button `Upvote'
    
    \quad\quad\quad\quad[5073] StaticText `1'
    
    \quad\quad\quad\quad[4581] button `Downvote'
    
    \quad\quad\quad[4310] article 

    \quad\quad\quad\quad[4313] HeaderAsNonLandmark 
    
    \quad\quad\quad\quad\quad[4585] heading```Star Trek: Strange New Worlds' (Season 3) and `Lower Decks' (Season 5) Renewed at Paramount+; `Strange New Worlds' Season 2 Premieres June 15, `Lower Decks' Season 4 Premieres Summer 2023"
    
    \quad\quad\quad\quad\quad\quad[5081] link ```Star Trek: Strange New Worlds' (Season 3) and `Lower Decks' (Season 5) Renewed at Paramount+; `Strange New Worlds' Season 2 Premieres June 15, `Lower Decks' Season 4 Premieres Summer 2023"
    
    \quad\quad\quad\quad[4591] button `Upvote'
    
    \quad\quad[4448] complementary 
    
    \quad\quad\quad[4475] HeaderAsNonLandmark 
    
    \quad\quad\quad\quad[4832] heading `MarvelsGrantMan136'
    
    \quad\quad\quad\quad\quad[5500] link `MarvelsGrantMan136'
    
    \quad\quad\quad\quad[5503] StaticText `Registered '
    
    \quad\quad\quad\quad[5504] time `June 24, 2020 at 11:02:10 AM EDT'
    
    \quad\quad\quad\quad\quad[6184] StaticText `6 years ago'
    
    \quad\quad\quad\quad[5507] StaticText `Not whitelisted'
    
    \quad\quad\quad[4835] StaticText `t2\_5adwlxvnAwesome Prompt ArtistFreelance Web DeveloperAwesome Prompt ArtistAwesome Prompt ArtistAwesome Prompt ArtistFreelance Web DeveloperI am a robotI am a robotPro Python Developer withFreelance Web Developer Freelance Web Developer Freelance Web Developer 20 years of Experience'
    
    \quad\quad\quad[4836] link `Edit biography'
    
    \quad\quad\quad[4478] heading `Toolbox'
    
    \quad\quad\quad[5512] link `Hidden forums'
    
    \quad\quad\quad[5513] link `Trash'
    
    \quad\quad\quad[4480] heading `MarvelsGrantMan136 is a moderator on:'
    
    \quad\quad\quad[5514] link `PlantsForCatParents'
    
    \quad\quad\quad[5515] link `cmu\_lti'
    
    \quad\quad\quad[5516] link `KaraokeKaraoke'
    
    \quad\quad\quad[5517] link `Karaoke'
    
    \quad\quad\quad[5518] link `sci\_fi'
    
    \quad\quad\quad[5519] link `cmu\_lticmu\_lti'
    
    URL: \url{http://3.151.53.169:9999/user/MarvelsGrantMan136/submissions}
}

\vspace{2mm}\noindent\tikz\draw[dashed, gray!50!black] (0,0) -- (\linewidth,0);\vspace{2mm}

\textbf{3. TaskExecutor's Reasoning and Action (\texttt{gpt-4o-mini}):} \\
Let's think step-by-step. According to the memory management agent, I have navigated to the comments section for the user `MarvelsGrantMan136'. The comments listed show that there are a total of 4 comments made by this user, and all of them either have no votes or upvotes, but none have received more downvotes than upvotes. Therefore, the count of comments that have received more downvotes than upvotes is 0. 
    
In summary, the next action I will perform is ```stop [0]'''

\vspace{2mm}\noindent\tikz\draw[dashed, gray!50!black] (0,0) -- (\linewidth,0);\vspace{2mm}

\textbf{4. Environment Feedback:} \\
The agent correctly deduces that the count is 0 and issues the `stop` action, indicating successful task completion and a correct final answer (reward 1).
\end{tcolorbox}

\subsection{Conclusion of Case Study}
This case study vividly demonstrates the \textit{ActiveContext} architecture's ability to effectively condense a verbose multi-turn interaction history into concise, actionable memory. This mechanism allows the \texttt{gpt-4o-mini} model to maintain focus on the complex distal objective, avoid Attention Dilution from massive DOM trees, and ultimately arrive at the correct answer efficiently. The dynamic refinement of working memory by the \textit{ContextCurator} is crucial for enabling robust performance in long-horizon tasks, contributing significantly to both token efficiency and reasoning accuracy.

\end{document}